\documentclass[aos]{imsart}

\RequirePackage{amsthm,amsmath,amsfonts,amssymb}
\RequirePackage[authoryear]{natbib}
\RequirePackage[colorlinks,citecolor=blue,urlcolor=blue]{hyperref}
\RequirePackage{graphicx}
\usepackage{subfigure}
\usepackage{float}
\usepackage{multirow}
\usepackage{diagbox}
\usepackage[figuresright]{rotating}
\usepackage{pdflscape}
\usepackage{bm}
\usepackage{algorithm, algorithmic}
\usepackage{enumerate}
\usepackage{booktabs}
\usepackage{mathabx} 
\startlocaldefs 
\theoremstyle{plain}

\newtheorem{theorem}{Theorem}[section]
\newtheorem{lemma}{Lemma}[section]
\newtheorem{remark}{Remark}[section]
\theoremstyle{remark}
\newtheorem{definition}{Definition}[section]

\newtheorem{assumption}{Assumption}
\newtheorem{corollary}{Corollary}[section]

\newcounter{cnstcnt}

\newcounter{bnstcnt}

\newcounter{dbnstcnt}

\makeatletter
\newenvironment{breakablealgorithm}
  {
   \begin{center}
     \refstepcounter{algorithm}
     \hrule height.8pt depth0pt \kern2pt
     \renewcommand{\caption}[2][\relax]{
       {\raggedright\textbf{\ALG@name~\thealgorithm} ##2\par}%
       \ifx\relax##1\relax 
         \addcontentsline{loa}{algorithm}{\protect\numberline{\thealgorithm}##2}%
       \else 
         \addcontentsline{loa}{algorithm}{\protect\numberline{\thealgorithm}##1}%
       \fi
       \kern2pt\hrule\kern2pt
     }
  }{
     \kern2pt\hrule\relax
   \end{center}
  }
\makeatother

\makeatletter
\newcommand*{\centerfloat}{%
  \parindent \z@
  \leftskip \z@ \@plus 1fil \@minus \marginparwidth
  \rightskip \leftskip
  \parfillskip \z@skip}
\makeatother

\usepackage{enumerate}

\newcommand{\M}{\bm{M}}
\newcommand{\bH}{\bm{H}}
\newcommand{\Y}{\bm{Y}}

\newcommand{\bOmega}{\bm{\Omega}}
\newcommand{\U}{\bm{U}}
\newcommand{\V}{\bm{V}}
\newcommand{\bu}{\bm{u}}
\newcommand{\bv}{\bm{v}}

\newcommand{\N}{\bm{N}}

\makeatletter
\newcommand*{\addFileDependency}[1]{
\typeout{(#1)}
\@addtofilelist{#1}
\IfFileExists{#1}{}{\typeout{No file #1.}}
}
\makeatother

\endlocaldefs

\begin{document}

\begin{frontmatter}
\title{Matrix completion via residual spectral matching}
\runtitle{Residual spectral matrix completion}

\begin{aug}


\author[A]{\fnms{Ziyuan}~\snm{Chen}\ead[label=e1]{chenziyuan@pku.edu.cn}},
\author[A]{\fnms{Fang}~\snm{Yao}\ead[label=e2]{fyao@math.pku.edu.cn}}


\address[A]{School of Mathematical Sciences, Center for Statistical Science, Peking University\printead[presep={,\\ }]{e1,e2}}

\end{aug}

\begin{abstract} 
Noisy matrix completion has attracted significant attention due to its applications in recommendation systems, signal processing and image restoration. 
Most existing works rely on (weighted) least squares methods under various low-rank constraints. 
However, minimizing the sum of squared residuals is not always efficient, as it may ignore the potential structural information in the residuals.
In this study, we propose a novel residual spectral matching criterion that incorporates not only the numerical but also locational information of residuals. This criterion is the first in noisy matrix completion to adopt the perspective of low-rank perturbation of random matrices and exploit the spectral properties of sparse random matrices.
We derive optimal statistical properties by analyzing the spectral properties of sparse random matrices and bounding the effects of low-rank perturbations and partial observations. 
Additionally, we propose algorithms that efficiently approximate solutions by constructing easily computable pseudo-gradients. 
The iterative process of the proposed algorithms ensures convergence at a rate consistent with the optimal statistical error bound.
Our method and algorithms demonstrate improved numerical performance in both simulated and real data examples, particularly in environments with high noise levels.
\end{abstract}



\end{frontmatter}


\section{Introduction}

\subsection{Literature review}
Matrix completion aims to reconstruct the target matrix $\M \in \mathbb{R}^{m \times n}$ from partial observations of its entries that are corrupted by random noise, which has applications across different fields.
For example, it has been utilized in recommendation systems \citep{jannach2016recommender,ramlatchan2018survey}, signal processing \citep{weng2012low,li2019survey}, and image restoration \citep{ji2011robust,jia2022non}.
Recently, matrix completion based statistical models have also been applied to various problems, including wireless sensor networks \citep{xie2016recover,kortas2021robust}, causal inference \citep{kallus2018causal,athey2021matrix}, and policy evaluation \citep{iweze2020matrix,duan2024online}. 

In the matrix completion model, the sample size is often significantly smaller than the matrix dimensions $m\times n$. Completing the target matrix $\M$ is generally impossible without specific structures and properties. The mostly considered property for the underlying matrix $\M$ is a low-rank structure or related characteristics, such as the rapid decay of singular values. A common approach to addressing the low-rank matrix completion problem is to impose strict low-rank constraints through specific matrix factorization techniques. For instance, \cite{keshavan2009matrix} introduced a method that employs thin singular value decomposition. The typical method for constraining the rank of matrix $\M$ to be less than $s$ represents it as $\M = \bm{L}\bm{R}'$, where $\bm{L}\in \mathbb{R}^{m\times s}$ and $\bm{R}\in \mathbb{R}^{n\times s}$. Although this low-rank representation lacks identifiability, extensive research has examined the properties of least squares methods utilizing this matrix factorization with alternative minimizing schemes. This includes the incorporation of regularization terms \citep{sun2016guaranteed,ge2016matrix} and algorithmic analyses \citep{jain2013low,hardt2014understanding,zheng2016convergence,ma2018implicit,chen2020nonconvex}. Following the findings of \cite{candes2009exact}, another prevalent method has emerged. They demonstrated that exact matrix completion is achievable in the absence of random noise, by relaxing the strict rank constraint to a nuclear norm constraint. Consequently, the nuclear norm constraint is widely used in both practical applications and theoretical analyses due to its convex properties. Various least squares estimators based on the nuclear norm \citep{candes2010matrix,rohde2011estimation,negahban2011estimation} and their variants \citep{koltchinskii2011nuclear,cai2015rop,cai2016matrix} have been thoroughly investigated.

Recent studies have also explored various extensions to traditional noisy matrix completion settings. Works such as \cite{klopp2014noisy} and \cite{bi2017group} extended the missing data mechanism from ``missing completely at random''  to ``missing at random''. 
Later, the missing mechanism was further extended to ``missing not at random'' \citep{bhattacharya2022matrix,choi2023matrix}.  \cite{chen2019inference} and \cite{xia2021statistical} investigated both completion and statistical inference problems. \cite{gui2023conformalized} later combined conformal inference techniques with matrix completion. Using additional covariate information, \cite{mao2019matrix} and \cite{jin2022matrix} proposed a column space decomposition model. \cite{ma2024statistical} also studied the inference problem related to this model. \cite{xu2016dynamic} and \cite{chen2024dynamic} introduced dynamic matrix estimators designed for time-varying target matrices. In the absence of low-rank structure, high rank matrix completion \citep{eriksson2012high,elhamifar2016high,ongie2017algebraic} and nonlinear variants \citep{fan2017deep,fan2018matrix} have also been extensively studied. Additionally, tensor completion has gained popularity due to the development of its efficient algorithms \citep{zhang2020islet,luo2023low}. 

\subsection{Different perspective from random matrix theory}
For the noisy matrix completion problem and its variants, previous works typically treats the noisy matrix completion model as data sampled from the target matrix $\M$ with random noise. The commonly used criterion for evaluating performance in noisy matrix completion is the (weighted) squared error. Least squares methods are often employed in statistical models involving noise, and the resulting estimators generally exhibit good performance. However, the rank structure of matrices relates to both the values and locations of entries, indicating that random noises possesses both numerical and locational characteristics in the noisy matrix completion problem. Applying least squares methods directly overlooks the locational properties of residuals and may not be efficient, especially in high-noise scenarios. 

In contrast, we approach the noisy matrix completion problem from a different perspective.
We model the data as samples from a low-rank perturbation $\M$ of a random matrix $\bH$ consisting of random noise. This perspective motivates us to propose a criterion based on the similarity between the residual matrix and sparse random matrices. Inspired by the fact that singular values capture both numerical and positional information about entries in matrices, we measure the similarity by comparing the distribution of singular values between the residual matrix and sparse random matrices.
The properties of singular values of mean zero rectangular random matrices $\bm{X}\in\mathbb{R}^{m\times n}$ with $\mathbb{E}[\bm{X}^2_{i,j}] = 1$, are directly linked to the empirical spectral distribution (ESD) of sample covariance matrices $\bm{S} = \bm{X}\bm{X}'/m$. This ESD converges to the Marchenko-Pastur (M-P) distribution as $m \rightarrow \infty$ with the ratio $n/m \rightarrow \rho > 0$ \citep{yin1986limit,bai1988necessary}. 
Extensive research has examined the convergence rate to the M-P distribution \citep{gotze2004rate,gotze2011rate,bai2012convergence}, local convergence in various settings \citep{bourgade2014local,alt2017local,bao2017local} and the universality property \citep{erdos2010bulk,erdHos2012rigidity,tao2012random,pillai2014universality}.
\cite{benaych2012singular} provided the limit values of leading singular values and vectors in low-rank perturbations of random matrices. Subsequent studies have explored additional properties, such as the limit distribution of leading singular values \citep{veraart2016denoising,bao2021singular,bao2022statistical}. The tail distributions of entries in classical random matrix theories are often constrained by bounded high-order moment conditions. 
Sparse random matrices, where each entry is multiplied by a random variable following a binomial distribution with probability $p_{i,j}\rightarrow 0$, do not satisfy these conditions. 
For example, the entries in Gaussian sparse random matrices $\bm{X}$ with $\mathbb{E}[\bm{X}^2_{i,j}] = 1$ have a divergent fourth moment $\mathbb{E}[\bm{X}^4_{i,j}] = 3p_{i,j}^{-1}$. 
Recently, the spectral properties of sparse random matrix have attracted much attention. For example, the local law for sparse sample covariance matrices has been established \citep{lee2021higher,gotze2022local,kafetzopoulos2023local}. The universality property of these matrices is also discussed in several studies \citep{ding2018necessary,he2023edge,brailovskaya2024universality}.

\subsection{Our contributions and paper organization}

In this paper, we propose a novel residual spectral matching criterion 
for noisy matrix completion. We adopt the perspective of low-rank perturbation of random matrices and leverage the properties of sparse random noise matrices to design this effective criterion. The criterion incorporates both the locational and numerical information of residuals, unlike the Frobenius norm loss which considers only the numerical aspect. To the best of our knowledge, this is the first approach that accounts for the location of residuals in noisy matrix completion.
The proposed criterion is formulated as the weighted $L_2$ distance between the ordered singular values of the residual matrix and the expected singular values of variance-adjusted sparse Gaussian random matrices.
Adjusting the spectral distribution of the residual matrix to resemble that of sparse random matrices provides more structural information than merely minimizing the Frobenius norm of the residual matrix. 
Our research demonstrates both optimal theoretical properties and improved numerical performance for this criterion, especially at a high noise level.
 
The main contributions of our study are four-fold. First, a new residual spectral matching criterion is proposed for noisy matrix completion. This criterion is user-friendly and adaptable to various low-rank constraints. Second, the statistical properties of estimators under both strict low-rank and relaxed nuclear norm constraints are provided. We develop an analytical framework that leverages techniques from low-rank perturbation, random missing data, and sparse random matrices. Oracle bounds are presented, and optimal upper bounds are achieved by selecting the appropriate rank constraint and tuning hyper-parameter for the nuclear norm constraint. Third, a general iterative algorithm for minimizing the proposed criterion is presented. An alternating descent scheme and a pseudo-gradient formula are introduced to handle the non-convexity of the criterion and the challenge of formulating explicit gradients. Finally, the properties of algorithm iterations are carefully analysed. We demonstrate that our algorithm achieves the optimal error bound within a finite number of iterations by monitoring both the incoherence condition and the distance between iterative series and the target matrix.

The structure of the paper is as follows. Section \ref{sec:preliminary} provides a brief introduction of the noisy matrix completion problem, and discusses the success of the low-rank perturbation viewpoint and the random matrix theory in the matrix denoising model, a special case of noisy matrix completion. In Section \ref{sec:methodology}, we present the spectral matching criterion, estimation methods and the corresponding algorithm. Section \ref{sec:theoretical} establishes the optimal statistical properties of the proposed estimators and the error analysis for the iterative series in the proposed algorithm. In Section \ref{sec:numerical}, both simulation and real data experiments demonstrate the favorable performance of the proposed methods over commonly used Frobenius norm methods, particularly when dealing with high noise levels. Conclusions and additional issues are discussed in Section \ref{sec:discussion}. Proof sketches and algorithms are presented in the Appendix, while additional technical lemmas and detailed proofs are provided to an online Supplementary Material for space economy.
\section{Preliminary and Motivation}
\label{sec:preliminary}

\subsection{Settings for noisy matrix completion and notations}

We first introduce some notations to be used throughout the paper. For any matrix $\M\in \mathbb{R}^{m\times n}$, the following norms are defined: $\|\M\|$ for the spectral norm (the largest singular value), $\|\M\|_{F}$ for the Frobenius norm (the square root of the sum of squared entries), $\|\M\|_{*}$ for the nuclear norm (the sum of singular values), $\|\M\|_{\infty}$ for the maximum norm (the largest absolute value of entries), and $\|\M\|_{2,\infty}$ for the largest $L_2$ norm of the rows (sometimes named as the $L_2/L_\infty$ norm \citep{ma2018implicit}). The transpose of $\M$ is denoted by $\M'$, and $\operatorname{Proj}_{s}(\M)$ indicates the projection of $\M$ onto the space of matrices with rank no more than $s$. We use the notation $\M_{[1:s]}$ to denote the matrix formed by the first $s$ columns of $\M$, and $(a)_{+}$ to represent the positive part of $a$. Additionally, the notations $a(m) \lesssim b(m)$ means that there exists a constant $c>0 $ such that $|a(m)|\le c|b(m)|$ and similarly $a(m) \gtrsim b(m) $ means that $|a(m)|\ge c|b(m)|$.

Noisy matrix completion involves recovering the target matrix $\M_0\in \mathbb{R}^{m\times n},n/m=\rho \in (0,1]$ from partial observations of its noisy perturbations. Formally speaking, it aims to estimate $\M_0$ from the observation $Y\in \mathbb{R}^{m\times n}$
\begin{align*}
\Y = P_{\bOmega}(\M_0 + \bH),
\end{align*}
where $\bOmega\in \mathbb{R}^{m\times n}$ is a binary matrix with ones indexing the locations of observed entries, and $\bH\in \mathbb{R}^{m\times n}$ is an unknown noise matrix that is independent of $\M_0$. The projection $P_{\bOmega}(\cdot)$ preserves the observed entries in $\bOmega$ and assigns zeros to the unobserved entries.

In previous related works, $\bOmega_{i,j}, 1\le i\le m,1\le j\le n$ are often assumed to follow independent binomial distributions with probability $p$ and $H_{i,j}, 1\le i\le m,1\le j\le n$ are independent mean zero random variables with variance $\sigma^2$. To separate the target matrix $\M_0$ from the random noise $\bH$, a key assumption is that $\M_0$ has a low rank $r$ which is far less than $n$. The incoherent condition defined in Definition \ref{def:incoherent} is also necessary for $\M_0$, to ensure that the missing pattern $\bOmega$ does not completely disrupt the singular subspace information of $\M_0$. 
\begin{definition}
\label{def:incoherent}
        A rank-$r$ matrix $\M\in \mathbb{R}^{m\times n}$ is said to be $\mu$-incoherent if 
    \begin{align*}
        \|\U\|_{2,\infty} \le \sqrt{\frac{\mu r}{m}}, \quad \|\V\|_{2,\infty} \le \sqrt{\frac{\mu r}{n}},
    \end{align*}
    where $\M = \U \bm{D} \V'$ is the singular value decomposition.
\end{definition}
This incoherent condition of a low-rank matrix $\M$ implies the delocalization of its singular vectors, ensuring that each entry of $\M$ contributes similarly important information to the matrix as a whole. This condition, in turn, provides favorable properties for projection operator $P_{\bOmega}(\cdot)$ onto sparse observations. 
Specifically, the restricted isometry property (RIP) $(1-\delta)\|\M\|_{F}^2 \le\|p^{-1/2}P_{\bOmega}(\M)\|_{F}^2 \le (1+\delta)\|\M\|_{F}^{2}$ holds for $P_{\bOmega}(\cdot)$ with high probability, where $\delta\ge 0$ is sufficiently small.

Two mainstream methods and their variations have been proposed and extensively studied for noisy matrix completion. The first type is mostly based on the low-rank factorization $\M = \bm{L}\bm{R}',\bm{L}\in \mathbb{R}^{m\times s},\bm{R}\in\mathbb{R}^{n\times s}$ of a rank $s$ matrix $\M$. The matrix $\M_0$ is then estimated by
\begin{align}
\label{eq:nonconvex-estimate}
    \widehat{\M}_{\text{fac}} = \underset{\M = \bm{L}\bm{R}',\bm{L}\in \mathbb{R}^{m\times s},\bm{R}\in\mathbb{R}^{n\times s}}{\arg \min}  \|\Y - P_{\bOmega}(\bm{L}\bm{R}')\|_{F}^2.
\end{align}
Several variations add different regularization terms, such as the Frobenius norm penalty $\lambda(\|\bm{L}\|_{F}^2 + \|\bm{R}\|_{F}^2)$ or the balanced penalty $\lambda\|\bm{L}'\bm{L} - \bm{R}'\bm{R}\|_{F}^2$. The second type constrains the rank of the estimator by controlling its nuclear norm, i.e., 
\begin{align}
\label{eq:convex-estimate}
    \widehat{\M}_{\text{nuc}} =  \underset{\M\in  \mathbb{R}^{m\times n}}{\arg \min}  \|\Y - P_{\bOmega}(\M)\|_{F}^2 +\lambda \|\M\|_{*}.
\end{align}
In the absence of noise matrix $\bH$, Theorem 1.7 in \cite{candes2010power} shows that the lower bound for $p$ to achieve exact matrix completion with high probability $1-\delta$ is $p \gtrsim \mu_0 r m^{-1}\log (m/2\delta)$. For sufficiently large $p$, the lower bound for the Frobenius norm distance between $\M_0$ and its estimator $\widehat{\M}$ is $\|\widehat{\M}-\M_0\|_{F} \gtrsim \sigma \sqrt{mr/p}$. Both estimators mentioned above have been proven to achieve the optimal or near-optimal (up to a logarithmic factor $\log m$) upper bounds. These results can be found in methods based on matrix factorization \citep[e.g.,][]{keshavan2009matrix,ma2018implicit,chen2020noisy} and  for convex relaxed methods \citep[e.g.,][]{koltchinskii2011nuclear,klopp2014noisy}.

\subsection{A motivating example: matrix denoising}

The matrix denoising problem aims to separate the underlying signal matrix $\M_0$ from its noisy version $\M_0 + \bH$. In fact, the model can be viewed as a special case of matrix completion with $p = 1$, where $\Y = \M_0 +\bH$ is fully observed. When we apply the estimation methods \eqref{eq:nonconvex-estimate} and \eqref{eq:convex-estimate} in parallel to the matrix denoising problem, two estimators for $\M_0$ are obtained respectively:
\begin{align*}
    \widecheck{\M}_{\text{fac}} = \sum_{i=1}^s \sigma_i \bu_i \bv_i',\quad \widecheck{\M}_{\text{nuc}} = \sum_{i=1}^n (\sigma_i - \lambda/2)_{+}\bu_i \bv_i',
\end{align*}
where $\Y = \sum_{i=1}^n \sigma_i \bu_i \bv_i'$ is the singular value decomposition of $\Y$. 
While both estimators achieve optimal upper bounds that match the problem's lower bound $\sigma \sqrt{rm}$, we shall demonstrate that neither performs the best in matrix denoising compared to the estimator based on random matrix theory. And the performance gap widens as the noise level rises. 

Rather than viewing the matrix denoising model as a target matrix contaminated by random noise, we consider the model as a random matrix $\bH$ perturbed by a low-rank matrix $\M_0$. Classical random matrix theory \citep{bai2010spectral} shows that the empirical spectral distribution of $\bH'\bH/m$ converges to the Marchenko-Pastur distribution. Consequently, the density of the singular values of $\bH/\sqrt{m}$ converges to
\begin{align*}
    \nu(x) = \frac{1}{\pi \sigma^2}\frac{\sqrt{(\lambda^2_{+}-x^2)(x^2-\lambda^2_{-})}}{\rho x}\mathbf{1}_{x\in[\lambda_{-},\lambda_{+}]},\quad \lambda_{\pm} = \sigma(1\pm\sqrt{\rho}).
\end{align*}
The smallest $s$ singular values of $\bH$ are found to be larger than $\sqrt{m}\lambda_{-}$ with high probability. Taking $\widecheck{\M}_{\text{fac}}$ as an example, $\widecheck{\bH} = \Y - \widecheck{\M}_{\text{fac}}$ can intuitively be viewed as an estimator of the random noise matrix $\bH$. However, the $s$ smallest singular values of $\widecheck{\bH}$ are zero, which differs from those of the random noise matrix $\bH$. This discrepancy indicates that $\widecheck{\M}_{\text{fac}}$ may not be adequate. \cite{benaych2012singular} derived the limit distribution of singular values of random matrices with low-rank perturbations. It helps us to construct the estimator $\widecheck{\M}_{\text{rmt}} = \sum_{i=1}^s \widetilde{\sigma}_i \bu_i \bv_i'$, where
\begin{align*}
    \widetilde{\sigma}_i = \left(\frac{1}{2}\left(\left(\sigma_i^2 - \widehat{\sigma}^2(1+\rho)m\right) +\sqrt{\left(\widehat{\sigma}^2(1+\rho)m -\sigma_i^2\right)^2-4\widehat{\sigma}^4\rho m^2}\right)\right)^{1/2},
\end{align*}
and $\widehat{\sigma}$ is a $\sqrt{m}$-consistency estimator of $\sigma$. To verify our motivation, We conduct a numerical study to compare the performance of the estimator $\widecheck{\M}_{\text{rmt}}$ with $\widecheck{\M}_{\text{fac}}$ and $\widecheck{\M}_{\text{nuc}}$ in Table \ref{tab:denoise_model}, which shows the outstanding performance of the estimator $\widecheck{\M}_{\text{rmt}}$ based on random matrix theory, and the increasing performance gap for larger $\sigma$.

\begin{table}[tb]
\vspace{-0.25cm}
    \caption{The Frobenius norm error ($\times 10^{-2}$) $\|\widecheck{\M}-M_0\|_{F}/(mn)^{-1/2}$ for $\widecheck{\M}_{\text{fac}},\widecheck{\M}_{\text{nuc}}$ and $\widecheck{\M}_{\text{rmt}}$ across varying noise levels, based on $100$ repeated experiments with matrix dimension $m=500,n=250$ and the rank set to $r$.}
    \label{tab:denoise_model}
    \begin{tabular}{cc |c c c c c c c c c}
		\toprule[1pt]
		\multicolumn{2}{c}{Noise level $\sigma$}&0.01 & 0.02&0.05& 0.08 & 0.1 & 0.2 &0.3 & 0.4 & 0.5  \\
    \midrule[1pt]
    $r=5$ &$\widecheck{M}_{\text{rmt}}$ & 0.1724 & 0.3453& 0.8626 & 1.3796 & 1.7244 & 3.4465 & 5.1645 & 6.8835 & 8.4847\\
    $r=5$ &$\widecheck{M}_{\text{fac}}$ & 0.1724 & 0.3453& 0.8630 & 1.3812&1.7275 & 3.4717 & 5.2503 & 7.0831 & 8.8648\\
    $r=5$ &$\widecheck{M}_{\text{nuc}}$ & 0.2699 & 0.5396& 1.3360& 2.1219& 2.6390 & 5.1299 & 7.4743 & 9.7042 & 11.654\\
    \hline
    \hline
    $r=10$ &$\widecheck{M}_{\text{rmt}}$ & 0.2431 & 0.4869& 1.2154 & 1.9459 & 2.4301& 4.8577 & 7.2784& 9.6842 & 12.078\\
    $r=10$ &$\widecheck{M}_{\text{fac}}$ & 0.2431& 0.4870& 1.2160&1.9480&2.4344 & 4.8916 & 7.3943 & 9.9552&12.604\\
    $r=10$ &$\widecheck{M}_{\text{nuc}}$ & 0.3597 & 0.7183& 1.7776& 2.8225& 3.5043 & 6.8005 &9.8827 & 12.782 & 15.474\\
    \hline
    \hline
    $r=20$ &$\widecheck{M}_{\text{rmt}}$ &0.3415  &  1.0264& 1.8807 & 2.9057 & 3.4155 &  6.9952& 10.402 & 13.600 &16.453 \\
    $r=20$ &$\widecheck{M}_{\text{fac}}$ & 0.3415 & 1.0266 & 1.8816 & 2.9090 & 3.4211 &  7.0432&10.560  & 13.953 &17.078 \\
    $r=20$ &$\widecheck{M}_{\text{nuc}}$ & 0.4907 & 1.4653 & 2.6650 &  4.0801& 4.7751 &  9.4669& 13.650& 17.346 &20.442 \\
    \bottomrule[1pt]
    	\end{tabular}
\vspace{-0.25cm}
\end{table}


While random matrix theory has achieved significant success in matrix denoising models, its application to noisy matrix completion remains underexplored. 
The strong performance of the estimator $\widecheck{\M}_{\text{rmt}}$ in matrix denoising prompted us to adopt the perspective of low-rank perturbations of random matrices. In the noisy matrix completion problem, we treat the observation $\Y=P_{\bOmega}(\M_0 +\bH)$ as a perturbation of the sparse random matrix $P_{\bOmega}(\bH)$. Thus, finding an optimal estimator $\hat{\M}$ of $\M_0$ corresponds to approximating $P_{\bOmega}(\bH)$ through the residual matrix $\Y - P_{\bOmega}(\hat{M})$. Unlike matrix denoising, there is no closed-form solution for $\hat{M}$ in noisy matrix completion. Due to limitation in sample size, approximating each entry of $P_{\bOmega}(\bH)$ is challenging. We leverage the spectral distributions of sparse random matrices, which are stable and converge rapidly to their limit distributions. By using the distance between spectral distributions as a criterion, we can assess whether the residual matrix belongs to the high probability set of sparse random matrices. This approach offers a novel perspective on the noisy matrix completion problem.
\section{Methodology and algorithm}
\label{sec:methodology}

\subsection{Proposed method}
A sparse Gaussian random matrix of size $m\times n$ and missing probability $1-p_{i,j} \ge 0, 1\le i\le m, 1\le j\le n$ can be formulated as
\begin{align*}
   \M\in\mathbb{R}^{m\times n}, \quad \M_{i,j} = \bm{\Phi}_{i,j} \bOmega_{i,j}, 1\le i\le m,1\le j\le n,
\end{align*}
where $\bm{\Phi}_{i,j}$ are independent Gaussian random variables with mean zero and variance $1/m$, and $\bOmega_{i,j}$ follow independent binomial distributions with success probability $p_{i,j}$.
Define the probability space $\mathcal{S}_{m,n,p} = \left(\mathbb{R}^{m\times n},\mathcal{B}_{m,n},\mathcal{P}_{m,n,p}\right)$, where $\mathcal{B}_{m,n}$ is the Borel $\sigma$-algebra of $\mathbb{R}^{m\times n}$, and the probability measure $\mathcal{P}_{m,n,p}$ has the density function
\begin{align*}
    f_{m,n,p}(\M) = \prod_{i=1}^m\prod_{j=1}^n \left[p_{i,j}\phi\left(\sqrt{m}\M_{i,j}\right)+ (1-p_{i,j})\delta(\M_{i,j})\right],
\end{align*}
where $\phi(x)$ is the density function of standard Gaussian distribution, and $\delta(x)$ is the Dirac delta function. 
We denote the singular values of $\M$ as $\lambda_{1}(\M)\ge \lambda_2(\M)\ge \cdots \ge \lambda_n(\M)$ and combine them into a vector $\bm{\lambda}(\M)$. 
  In the probability space $\mathcal{S}_{m,n,p}$, we define the expectation of the random vector $\bm{\lambda}(\M)$ as
  $\bm{\lambda} = \mathbb{E}_{\M} \bm{\lambda}(\M)$. Although deriving the explicit formulation of $\bm{\lambda}$ is challenging, it can be estimated using a simple Monte Carlo procedure $\widehat{\bm{\lambda}} = \sum_{i=1}^\ell \bm{\lambda}(\M_i)/\ell$ where $\M_i,1\le i\le \ell$ are independent sparse Gaussian random matrices. The entries of $\widehat{\bm{\lambda}}$ are denoted as $\widehat{\lambda}_i,1\le i\le n$.
  

We then construct a criterion $l(\bm{N};\bm{\omega})$ to evaluate whether a matrix $\bm{N}\in \mathbb{R}^{m\times n}$ belongs to a high probability set
in $\mathcal{S}_{m,n,p}$. Let the ordered singular values of $\bm{N}$ be denoted as $\sigma_1(\bm{N})\ge \sigma_2(\bm{N})\ge \cdots\ge \sigma_n(\bm{N})$, and define the vector $\bm{\sigma}(\bm{N}) = (\sigma_1(\bm{N}),\cdots,\sigma_n(\bm{N}))$. The criterion calculates the weighted $L_2$ distance between $\bm{\sigma}(\bm{N})$ and $\widehat{\bm{\lambda}}$, i.e., 
$$l(\bm{N};\bm{\omega}) = \left\|\bm{\sigma}(\bm{N}) - \widehat{\bm{\lambda}}\right\|_{\bm{\omega}}^2 = \sum_{i=1}^n \omega_i \left(\sigma_i(\bm{N}) - \widehat{\lambda}_i \right)^2,$$
where $\bm{\omega} = (\omega_1,\cdots,\omega_n)$ is the weight satisfying $\sum_{i=1}^n \omega_i = 1$.

For noisy matrix completion, any matrix $\M\in \mathbb{R}^{m\times n}$ close to $\M_0$ will result in the residual matrix 
$$\Y - P_{\bOmega}(\M) = P_{\bOmega}(\M_0 - \M) + P_{\bOmega}(\bH)$$
 being close to $P_{\bOmega}(\bH)$. The latter is located in a high probability set of $\mathcal{S}_{m,n,p}$ after variance adjustment. 
Therefore, we propose the following method to find an estimator $\widetilde{\M}\in \mathbb{M}$ for $\M_0$ as
\begin{align}
\label{eq:general_estimation}
    \widetilde{\M} = \underset{\M\in \mathbb{M}}{\arg\min}\ l\left(\frac{\Y - P_{\bOmega}(\M)}{\sqrt{m} \widehat{\sigma}(\M)};\bm{\omega}\right).
\end{align}
Here $\widehat{\sigma}(\M)$ is the estimated working standard deviation, defined as 
\begin{align} 
\label{eq:working_variance}
    \widehat{\sigma}(\M) = \frac{\sum_{i=n/3}^{i=2n/3} \sigma_i\left(\Y - P_{\bOmega}(\M)\right)}{\sqrt{m}\sum_{i=n/3}^{2n/3} \lambda_i},
\end{align}
where $\sigma_i\left(\Y - P_{\bOmega}(\M)\right),1\le i\le n$ are the ordered singular values of $\Y - P_{\bOmega}(\M)$. The working estimator $\widehat{\sigma}(\M)$ approximates the standard deviation $\sigma$ of the noise by calculating the proportion of the bulk singular values' sum from the residual matrix and sparse random matrices. This approach is efficient because the perturbation $P_{\bOmega}(\M_0-\M)$ has less impart on the bulk singular values than on the edge singular values of $P_{\bOmega}(\bH)$.

When the estimation space $\mathbb{M}$ is chosen as $\mathbb{M}_s = \left\{\M\ |\ \M =\bm{L}\bm{R}',\bm{L}\in \mathbb{R}^{m\times s}, \bm{R}\in \mathbb{R}^{n\times s}\right\}$, it results in the matrix factorization based estimator $\widetilde{\M}_{\text{fac}}$
\begin{align}
\label{eq:our_estimator_nonconvex}
    \widetilde{\M}_{\text{fac}} = \underset{\M = \bm{L}\bm{R}',\bm{L}\in \mathbb{R}^{m\times s},\bm{R}\in \mathbb{R}^{n \times s}}{\arg\min} l\left(\frac{\Y - P_{\bOmega}(\bm{L}\bm{R}')}{\sqrt{m} \widehat{\sigma}(\bm{L}\bm{R}')};\bm{\omega}\right).
\end{align}
If the space $\mathbb{M}$ is defined as $\mathbb{M}_{C} = \left\{\M\in \mathbb{R}^{m\times n}\ | \ \|\M\|_{*}\le C\right\}$, we equivalently obtain the convex relaxed nuclear norm estimator $\widetilde{\M}_{\text{nuc}}$
\begin{align}
\label{eq:our_estimator_convex}
    \widetilde{\M}_{\text{nuc}} = \underset{M\in \mathbb{R}^{m\times n}}{\arg\min} \ l\left(\frac{\Y - P_{\bOmega}(\M)}{\sqrt{m}\widehat{\sigma}(\M)};\bm{\omega}\right) +\lambda\|\M\|_{*}.
\end{align}

It is notable that our proposal, unlike the Frobenius norm based methods, incorporates the spectral properties of the noise matrix rather than solely minimizing the Frobenius norm of the residual matrix. In many statistical problems, random noise lacks meaningful structural information, and minimizing the sum of squared errors often yields optimal results. However, in models involving a fixed low-rank matrix plus a random noise matrix, the low-rank matrix perturbs only a few dominant singular values of the random matrix significantly. Thus, the spectral distribution of the residual matrix in noisy matrix completion can provide additional insights. Particularly, in scenarios where the Signal-to-Noise Ratio (SNR) $\|\M_0\|_{F}/\|\bH\|_{F}$ is low, disregarding the information encoded in $\bH$ and merely reducing the norm of the residual matrix may result in a loss of efficiency.

\subsection{Optimal algorithm}
Minimizing the term $\|\Y - P_{\bOmega}(\M)\|_{F}^2$ in equations \eqref{eq:nonconvex-estimate} and \eqref{eq:convex-estimate} is straightforward due to its strong convexity. However, in our proposed method, minimizing $l((\Y - P_{\bOmega}(\M))/\sqrt{m}\widehat{\sigma}(\M);\bm{\omega})$ is more challenging because it is non-convex and lacks of a computable gradient. To address this, we have developed efficient algorithms and their convergence properties will be demonstrated in the next section. 

In our algorithms, $\M$ and $\widehat{\sigma}(\M)$ are updated alternatively in each iteration step. 
When $p_{i,j}=1$, $P_{\bOmega}(\M) = \M$ and the gradient for $\M$ of can be directly calculated as
\begin{align}
\label{eq:gradient_for_denoising}
    \nabla_{\M}l\left(\frac{\Y - \M}{\sqrt{m} \widehat{\sigma}};\bm{\omega}\right) =\frac{1}{\sqrt{m}\widehat{\sigma}} \sum_{i=1}^n \omega_i \left(\sigma_i(\M) - \sqrt{m}\widehat{\sigma}\widehat{\lambda}_i \right)\bu_i(\M) \bv'_i(\M),
\end{align}
where $\bu_i(\M),\bv_i(\M)$ are the corresponding left and right singular vectors of $\Y-P_{\bOmega}(\M)$. In the context of matrix completion where $p_{i,j}<1$, computing the gradient becomes non-trivial. A pivotal step in our algorithm is finding a direction that closely approximates the true gradient for iterative gradient descent. Inspired by the form of gradient in \eqref{eq:gradient_for_denoising}, we define the pseudo-gradient as
\begin{align}
\label{eq:gradient_for_our}
    \widehat{\nabla}_{\M} l\left(\frac{\Y - P_{\bOmega}(\M)}{\sqrt{m}\widehat{\sigma}};\bm{\omega}\right) = P_{\bOmega}\left(\frac{1}{\sqrt{m}\widehat{\sigma}} \sum_{i=1}^n \omega_i \left(\sigma_i(\M) - \sqrt{m}\widehat{\sigma}\widehat{\lambda}_i \right)\bu_i(\M) \bv'_i(\M)\right).
\end{align}
It is straightforward to verify that the pseudo-gradient coincides with the gradient when $p=1$, and we will later demonstrate its adequacy in our algorithms for $p < 1$. We provide the algorithm for the estimator $\widetilde{\M}_{\text{fac}}$ in Algorithm \ref{alg:nonconvex} and defer the algorithm for the estimator $\widetilde{\M}_{\text{nuc}}$ to the Appendix. 
\begin{breakablealgorithm}
  \renewcommand{\algorithmicrequire}{\textbf{Input:}}
  \renewcommand{\algorithmicensure}{\textbf{Output:}}
  \caption{Iteration Algorithm for $\widetilde{\M}_{\text{fac}}$}\label{alg:nonconvex}
  \begin{algorithmic}[1]
  \REQUIRE Observed data $\Y\in \mathbb{R}^{m\times n}$, missing index matrix $\bOmega\in \mathbb{R}^{m\times n}$ and used rank $s$.
  \ENSURE Completion matrix $\M$.
  \STATE $\U^{(0)},\bm{D}^{(0)},\V^{(0)} \leftarrow \operatorname{svd}(p^{-1}Y)$;
  \STATE $\bm{R}^{(0)} \leftarrow \U^{(0)}_{[1:s]}(\bm{D}^{(0)}_{[1:s]})^{1/2}$ and $\bm{L}^{(0)} \leftarrow \V^{(0)}_{[1:s]}(\bm{D}^{(0)}_{[1:s]})^{1/2}$;
  \STATE Sample $l$ standard Gaussian random matrix $\M_i\in \mathbb{R}^{m\times n}$ with singular value vector $\bm{\lambda}(\M_i),1\le i\le l$;
  \STATE $\widehat{\bm{\lambda}} \leftarrow \sum_{i=1}^l \bm{\lambda}(M_i)/l$, $\operatorname{part\_sum} \leftarrow \sum_{i=n/3}^{2n/3} \widehat{\lambda}_i$;
  \WHILE {$k\le K$}
  \STATE $\U^{(k+1)},\bm{D}^{(k+1)},\V^{(k+1)}\leftarrow \operatorname{svd}(\Y - P_{\bOmega}(\bm{R}^{(k)}\bm{L}^{(k)'})$;
  \STATE $\widehat{\sigma}^{(k+1)} \leftarrow \sum_{i=n/3}^{2n/3}\bm{D}^{(k+1)}_{i,i}/\operatorname{part\_sum}$;
  \STATE $\widehat{\nabla}^{(k+1)}_{\M} \leftarrow \sum_{i=1}^n w_i(\bm{D}^{(k+1)}_{i,i} -  \widehat{\sigma}^{(k+1)}\widehat{\lambda}_i)\U^{(k+1)}_{[i]}\V^{(k+1)'}_{[i]}$;
  \STATE $\bm{R}^{(k+1)} \leftarrow \bm{R}^{(k)} - \eta_{k+1} P_{\bOmega}(\widehat{\nabla}_{\M}^{(k+1)}) \bm{L}^{(k)}$;
  \STATE $\bm{L}^{(k+1)} \leftarrow \bm{L}^{(k)} - \eta_{k+1} P_{\bOmega}(\widehat{\nabla}_{\M}^{(k+1)}) \bm{R}^{(k)}$;
  \ENDWHILE
  \RETURN $\M = \bm{R}^{(K+1)}\bm{L}^{(K+1)'}$.
  \end{algorithmic}  
  \end{breakablealgorithm}

\section{Theoretical results}
\label{sec:theoretical}
Several common assumptions prevalent in noisy matrix completion researches are outlined here.
\begin{assumption}
    \label{assum:noise}
    The random noise matrix $\bH\in \mathbb{R}^{m\times n}$ satisfies that
    \begin{enumerate}[(a)]
        \item $\bH_{i,j},\ 1\le i\le m,1\le j\le n$ are independent random variables with mean zero and variance $\sigma^2$. 
        \item $\bH_{i,j},\ 1\le i\le m,1\le j\le n$ have sub-exponential tailed distribution, i.e., there exist constant $c>0$ such that
        \begin{align*}
            P\left(|\bH_{i,j}| \ge t\right) \le \exp\left\{-\frac{t}{c\sigma}\right\},\quad \forall t>T>0.
        \end{align*}
        \item \label{assum:1.c.} The standard deviation $\sigma$ is large enough such that 
        \begin{align*}
            \sigma \gtrsim \left((mn)^{-1/2}\|\M_0\|_{F}\right) r^{1/2}m^{-1/2} \log^{1/2}n.
        \end{align*}
    \end{enumerate}
\end{assumption}

\begin{assumption}
\label{assum:target}
The underlying fixed matrix $\M_0\in \mathbb{R}^{m\times n},n/m=\rho\le 1$, with a rank $r\ll n$ and singular value decomposition $\M_0 =\sum_{i=1}^r \sigma_i \bu_i \bv_i'$, satisfies that
    \begin{enumerate}[(a)]
        \item The condition number $\kappa = \sigma_1/\sigma_r$ is bounded by a constant,
        \item The rank $r$ matrix $\M_0$ is $\mu_0$-incoherent.
    \end{enumerate}
\end{assumption}

\begin{assumption}
    \label{assum:missing}
    Data are missing completely at random with observed probability $p_{i,j} = p, 1\le i\le m,1\le j\le n$ and $1\ge p \gtrsim \mu_0 rm^{-1}\log^{\alpha} m$ for some large enough constant $\alpha>0$.
\end{assumption}

Assumptions \ref{assum:noise}, \ref{assum:target}, and \ref{assum:missing} are standard for noisy matrix completion, except for \eqref{assum:1.c.} in Assumption \ref{assum:noise}. Assumption \ref{assum:noise}.\eqref{assum:1.c.} ensures the noise level is sufficiently high to make the structural information of the noise matrix $\bH$ useful. This mild assumption applies to a wide range of noise levels, approximately when $\sigma \ge m^{-1}n^{-1/2}\|\M\|_{F}$. Assumptions \ref{assum:missing} can be relaxed to include a missing at random mechanism, where $p_{i,j}$ are not all equivalent and may depend on $i$. We assume Assumptions \ref{assum:noise}, \ref{assum:target}, and \ref{assum:missing} hold throughout this section.

\subsection{Statistical properties of matrix factorization estimation}
We consider the oracle properties of $\widetilde{\M}_{\text{fac}}$. This involves finding a matrix $\M$ with rank at most $s$ that minimizes the loss function 
\begin{align*}
    \mathcal{L}(\M;\Y,\bOmega)  = p^{-1}l\left(\frac{\Y - P_{\bOmega}(\M)}{\sqrt{m}\widehat{\sigma}(\M)};\bm{\omega}\right).
\end{align*}
We verify that the loss function $\mathcal{L}(\M;\Y,\bOmega)$ is reasonable by showing that $\mathcal{L}(\M_0;\Y,\bOmega)$ is sufficiently small. Lemma \ref{lem:loss_for_true} asserts that at the point $\M_0$, the value of this loss function rapidly converges to zero with high probability.
\begin{lemma}
    \label{lem:loss_for_true} 
    With probability larger than $1-cn^{-3}$, it holds that
    \begin{align*}
        \mathcal{L}(\M_0;\Y,\bOmega) \le C_0 \log^{-2} m,
    \end{align*}
    with some constant $C_0\ge 0$ independent of $m,n,p,\sigma$ and $\M_0$.
\end{lemma}

In Assumption \ref{assum:noise}, we do not assume a specific distribution for the entries of the noise matrix $\bH$. Classical random matrix theory suggests that the spectrum of $\bH$ can be approximated by that of Gaussian random matrices. This property, known as universality, helps in determining suitable criteria for the spectrum of $\bH$. However, a different scenario arises when considering the sparse random matrix $P_{\bOmega}(\bH)$. A key technique for proving Lemma \ref{lem:loss_for_true} is to establish a similar universality property of sparse random matrices.

Define the estimation space
\begin{align*}
    \mathbb{M}_{s,\mu_1} = \left\{\M\in \mathbb{R}^{m\times n}\ |\ \operatorname{rank}(\M)\le s, \text{$\M$ is $\mu_1$-incoherent} \right\}.
\end{align*}
Utilizing the matrix factorization $\M=\bm{L}\bm{R}'$, we can only partially fulfill the conditions such that $\operatorname{rank}(\M)\leq s$ constrained in $\mathbb{M}_{s,\mu_1}$. Although no constraint is explicitly imposed to ensure that $\M$ adheres to any incoherent condition, we will demonstrate in Section \ref{sec:4.3} that the incoherent property is maintained implicitly throughout the iteration process. Therefore, we consider the properties of estimators constrained in the estimation space $\mathbb{M}_{s,\mu_1}$.

Firstly, we focus on the scenario where the rank $s$ of the estimation space $\mathbb{M}_{s,\mu_1}$ is properly selected, i,e., $s = r$. Consider the estimator
\begin{align*}
    \widetilde{\M} = \underset{\M\in \mathbb{M}_{r,\mu_1}}{\arg\min}\  \mathcal{L}(\M;\Y,\bOmega).
\end{align*}
We provide a lemma demonstrating that any sub-optimal estimator $\M$ within the space $\mathbb{M}_{r,\mu_1}$ will result in a sufficiently high value for the loss function $\mathcal{L}(\M;\Y,\bOmega)$.
\begin{lemma}
\label{lem:large_loss_to_large_error}
    For $\M\in \mathbb{M}_{r,\mu_1}$, if  
    \begin{align*}
        \|\M_0 - \M\|_F - (3+5\kappa)(2+2\sqrt{\rho}+\mathcal{C}^{1/2})\sigma\sqrt{mr/p}\gtrsim (\log^{-1} m) \sigma\sqrt{mr/p}.
    \end{align*} 
    then with probability larger than $1-cn^{-3}$,
    \begin{align*}
        \mathcal{L}(\M;\Y,\bOmega) > \mathcal{C}\omega_1.
    \end{align*}
\end{lemma}

Lemma \ref{lem:loss_for_true} demonstrates that at the point $\M_0$, the loss function yields a sufficiently small value. In contrast, Lemma \ref{lem:large_loss_to_large_error} illustrates that an inadequate estimator $\M$ results in a significantly large value of the loss function. Integrating these two findings, it is evident that there exists an $\M\in \mathbb{M}_{r,\mu_1}$ for which the loss $\mathcal{L}(\M;\Y,\bOmega)$ is small. This minimized loss ensures a small estimation error $\|\M_0 -\M\|_{F}$, leading to Theorem \ref{thm:upper_bound_for_rankr} that gives the upper bound of the Frobenius norm error $\|\widetilde{\M} - \M_0\|_{F}$ with high probability.
\begin{theorem}
    \label{thm:upper_bound_for_rankr}
    When the weight $\bm{\omega}$ satisfies that $\omega_1 \gg \log^{-2} m$, with probability larger than $1-cn^{-3}$, 
    \begin{align*}
        \|\widetilde{\M} - \M_0\|_{F} \le (3+5\kappa)(2+2\sqrt{\rho}+o(1))\sigma\sqrt{mr/p}.
    \end{align*}
\end{theorem}
\begin{remark}
    The upper bound presented in Theorem \ref{thm:upper_bound_for_rankr} coincides with the minmax lower bound as shown in Theorem 3 in \cite{negahban2012restricted} and Theorem 5 in \cite{koltchinskii2011nuclear}, with an order of $\sigma\sqrt{mr/p}$. This is owing to the ratio $\rho = n/m \le 1$, and we assume that the condition number $\kappa$ is bounded by a constant. It has been demonstrated that methods minimizing the squared Frobenius norm of the residual matrix $\Y - P_{\bOmega}(\M)$ also achieve optimal bounds. Consequently, our bound aligns with results of those methods. Additionally, we provide an explicit term in the upper bound of Theorem \ref{thm:upper_bound_for_rankr}, elucidating the influence of $\kappa$ and $\rho$ especially when they are no longer constant and increase as $m\rightarrow\infty$.
\end{remark}

Our method primarily relies on the spectral properties of the random noise matrix, which implies that the noise level cannot be too low. In Assumption \ref{assum:noise}.\eqref{assum:1.c.}, we impose a constraint on the lower bound of $\sigma$. This can be also viewed as constraining the Signal-to-Noise Ratio (SNR) $\|\M_0\|_{F}/\|\bH\|_{F}$, ensuring it is not too large. As previously discussed, this assumption holds for a wide range of noise levels, requiring only that the standard deviation $\sigma$ of the noise exceeds approximately $m^{-1}n^{-1/2}\|\M_0\|_{F}$. Even if this mild assumption is not satisfied, we can still bound the estimation error. The error bound will be affected by the Frobenius norm of $\M_0$ if \eqref{assum:1.c.} in Assumption \ref{assum:noise} is not satisfied, as demonstrated in Corollary \ref{cor:large_M0}.
\begin{corollary}
    \label{cor:large_M0}
    If Assumption \ref{assum:noise}.\eqref{assum:1.c.} does not hold, when the weight $\bm{\omega}$ satisfies that $\omega_1 \gg \log^{-2} m$, with probability larger than $1-cn^{-3}$, 
    \begin{align*}
        \|\widetilde{\M} - \M_0 \|_{F} \le (3+5\kappa)(2+2\sqrt{\rho}+o(1))\sigma\sqrt{\frac{mr}{p}}+ 8(3+5\kappa)(\mu_0+\mu_1)\sqrt{\frac{r^{3}\log n}{mnp}}\|\M_0\|_{F}.
    \end{align*}
\end{corollary}
\begin{remark}
    The additional term in Corollary \ref{cor:large_M0} becomes dominant when the standard deviation $\sigma$ of the noise is very small. In the extreme case of matrix completion without noise, i.e., $\sigma = 0$, Theorem 1.2 in \cite{candes2010power} demonstrates that exact completion can be achieved with high probability using \eqref{eq:convex-estimate}. In scenarios with minimal noise, uncertainty in the spectral properties of the noise matrix $\bH$ can obscure its structural information, reducing the effectiveness of our method. Directly minimizing the Frobenius norm of the residual matrix $\Y - P_{\bOmega}(\M)$ is more effective in this case. However, when the noise level is high, which is the more challenging scenario, the advantage of our method becomes evident.
\end{remark}

Several studies have investigated the estimation and testing issues concerning the underlying rank $r$ of $\M_0$, e.g., \cite{zhang2019statistical} and \cite{li2023robust}.
When the noise level is significant, precisely estimating the true rank of $M_0$ may not always be feasible. Specifically, there is uncertainty about whether the selected low-rank space $\mathbb{M}_{s,\mu_1}$ matches $\mathbb{M}_{r,\mu_1}$. Misjudging the underlying rank of $\M_0$ results in increased estimation errors. Nevertheless, for $s\ge r$ that is not excessively large, we establish that the Frobenius norm error of the estimator $\widetilde{\M}_s$, defined as
\begin{align*}
    \widetilde{\M}_{s} = \underset{\M\in \mathbb{\M}_{s,\mu_1}}{\arg\min} \mathcal{L}(\M;\Y,\bOmega)
\end{align*}
can still be controlled at a low level.
\begin{corollary}
    \label{cor:mis_ranks}
     When $r\le s \le mp \log^{-2} m$, the weight $\bm{\omega}$ satisfies that $\omega_1 \gg \log^{-2} m$ and the noise standard deviation $\sigma \ge s^{3/2}r^{-1}m^{-1}n^{-1/2}\log^{1/2} n \|\M_0\|_{F}$, with probability larger than $1-cn^{-3}$, 
    \begin{align*}
        \|\widetilde{\M}_s- \M_0\|_{F} \le  \left((2+4\kappa)r^{1/2} +(1+\kappa)s^{1/2}\right)(2+2\sqrt{\rho}+o(1))\sigma \sqrt{m/p}.
    \end{align*}
\end{corollary}
   When $s$ is selected as $s=r$, the upper bound stated in Corollary \ref{cor:mis_ranks} aligns with the results of Theorem \ref{thm:upper_bound_for_rankr}. If $s>r$, the estimation error $\|\widetilde{\M}_s - \M\|_{F}$ is bounded by the order $\sigma \sqrt{sm/p}$ when both $\kappa$ and $\rho$ are considered constants. A higher value of $s$ corresponds to a larger estimation space $\mathbb{M}_{s,\mu_1}$, leading to increased errors. An additional source of error arises from including the top singular subspace of the noise matrix in the estimator $\widetilde{\M}_s$, due to the excess rank $s > r$. Frobenius norm based methods tend to capture the entire top subspace of noise when minimizing the Frobenius norm of the residual matrix. In contrast, our proposed method is less affected because it also considers the spectral distribution of the random noise matrix.

\subsection{Statistical properties of convex relaxation estimation}
In addition to the strictly low-rank constraint $\M = \bm{L}\bm{R}'$, several studies such as  \cite{candes2009exact} suggest that the nuclear norm $\|\M\|_{*}$ serves as an effective convex relaxation alternative for the exact rank. Our method, which aims to minimize the spectral distance between the residual matrix and sparse random matrices, can employ a similar convex relaxation procedure for low-rank constraints. Formally, we investigate the properties of the estimator $\widetilde{\M}_\lambda$, defined as
\begin{align*}
    \widetilde{\M}_{\lambda} = \underset{\M\in\mathbb{R}^{m\times n}}{\arg\min} \mathcal{L}(\M;\Y,\bOmega) +\lambda \|\M\|_{*}.  
\end{align*}
Methods that minimizing the Frobenius norm benefit from the convex relaxation, as the overall estimation problem is strongly convex. This property enhances both theoretical analysis and computational feasibility. 
Previous techniques demonstrating the performance guarantee of $\widehat{\M}_{\text{nuc}}$ often relies on the convexity of the squared Frobenius norm and the nuclear norm. In contrast, analyzing our method is more challenging due to the non-convexity of $\mathcal{L}(\M;\Y,\bOmega)$. When focusing on estimation properties, the estimation space $\mathbb{M}_C = \left\{\M\in \mathbb{R}^{m\times n} \ | \ \|\M\|_{*} \le C\right\}$ is more difficult to control compared to $\mathbb{M}_{s,\mu_1}$. This difficulty arises from the ambiguity surrounding the rank of matrices in $\mathbb{M}_C$. The following theorem provides an error bound for the estimator $\widetilde{\M}_{\lambda}$ with high probability for any $\lambda \ge 0$. 
\begin{theorem}
    \label{thm:upper_bound_for_convexlambda}
   For some $\bar{\omega} >0$, define that 
\begin{align*}
    R_1 = 2(1+\sqrt{\rho})\sigma \sqrt{m/p},\quad R_2 =  
    R_1+ \frac{\lambda  \sigma^2 m}{\bar{\omega} p}+ \left(\frac{\lambda \sigma^2 m}{\bar{\omega} p}\right)^{1/2}R_1^{1/2}.
\end{align*}
If $\min_{1\le i\le r(R_1+R_2)/R_1}\omega_i \ge  \bar{\omega}$,
with probability larger than $1-cn^{-3}$,
\begin{align*}
   \|\widetilde{\M}_{\lambda} - \M_0\|_{F} &\le (4+5\kappa) \left(\left(\frac{\lambda \sigma^2 m}{\bar{\omega} p}\right)^{1/2}\left(rR_2 - R_1\right)^{1/2} +R_1 + R_2\right) \\&\quad\quad\quad\quad\quad\quad\quad\quad\quad\quad+ \left((C_0+1)\lambda^{-1}\log^{-2} m + 2rR_2\right)^{1/2}R_1^{1/2}.
\end{align*}
\end{theorem}
The upper bound in Theorem \ref{thm:upper_bound_for_convexlambda} illustrates two different sources of error related to the choice of the tuning parameter $\lambda$. One source is the term $(C_0+1)\lambda^{-1}\log^{-2}m $, which arises from the flexibility of the estimation space. A higher $\lambda$ reduces this flexibility, thereby decreasing the generalization error. The other source is the term $\lambda \sigma^2 m/\bar{\omega} p$, which results from the shrinkage of singular values due to nuclear norm regularization. A smaller $\lambda$ leads to decreased approximation error. These two distinct sources of error create a trade-off when tuning $\lambda$ to minimize the bound, as illustrated in Corollary \ref{cor:convex_tuning_lambda}.
\begin{corollary}
    \label{cor:convex_tuning_lambda}
    When the weight $\bm{\omega}$ satisfies that $\min_{1\le i \le 4r} \omega_i \gg \log^{-2} m$, choosing $\lambda =  2(1+\sqrt{\rho}) \sigma^{-1}p^{1/2}m^{-1/2}\log^{-2} m$, with probability larger than $1-c n^{-3}$,
    \begin{align*}
        \|\widetilde{\M}_{\lambda} - \M_0\|_{F} \le 
        2(1+\sqrt{\rho})\left((\sqrt{6} + 4\sqrt{3}+5\sqrt{3}\kappa )r^{1/2}+4(4+5\kappa) +o(1)\right)\sigma\sqrt{m /p}.
    \end{align*}
\end{corollary}
 By adjusting the tuning parameter $\lambda$, our results in Corollary \ref{cor:convex_tuning_lambda} match the minmax lower bound for noisy matrix completion. Estimation properties for methods based on minimizing the squared Frobenius norm of the residual matrix with nuclear norm regularization, have been well understood. For example, assuming $\M_0$ is a $\mu$-incoherent matrix, Corollary 2 in \cite{koltchinskii2011nuclear} provides a nearly optimal upper bound of the order $\sigma \sqrt{m r\log m /p}$. Compared to results from classical convex relaxation methods, our upper bound no longer contains logarithmic factors.
\subsection{Algorithmic properties of iterative procedure}
\label{sec:4.3}
Iterative gradient descent algorithms are proposed to find approximate solutions, as closed-form solutions for the proposed methods are not feasible. Due to the non-convex nature of the loss function $\mathcal{L}(\M;\Y,\bOmega)$ and the non-convex constraint $\operatorname{rank}(\M)\le s$, general optimization theory does not guarantee global convergence. It is essential to accurately measure the performance gap between estimators (often the global minimizer) and the results obtained through iterative algorithms. For the constraint $\operatorname{rank}(\M) \le s$, methodologies such as matrix factorization $\M = \bm{L}\bm{R}'$ with alternating descent schemes, have been comprehensively researched. Therefore, our primary focus is on addressing the challenges arising from the proposed loss function $\mathcal{L}(\M;\Y,\bOmega)$. 

We formally define the iterative procedure for proposed method $\min_{\M\in \mathbb{M}_{s,\mu_1}} \mathcal{L}(\M;\Y,\bOmega)$. With a feasible initial value $\M^{(0)}$, we define the iterative series $\left\{\M^{(k)}\right\}_{k=0,1,\cdots}$ and the nuisance series $\left\{\N^{(k)}\right\}_{k=1,2,\cdots}$ as
\begin{align*}
    \N^{(k+1)} &= \M^{(k)} - \eta_{k+1} \widehat{\nabla} \mathcal{L}(\M^{(k)};\Y,\bOmega),
    \\ \M^{(k+1)} &= \operatorname{Proj}_{s}(\N^{(k+1)}), \quad k=0,1,\cdots,
\end{align*}
where 
$$ \widehat{\nabla} \mathcal{L}(\M;\Y,\bOmega) = p^{-1}\widehat{\nabla}_{\M} l\left(\frac{\Y - P_{\bOmega}(\M)}{\sqrt{m}\widehat{\sigma}};\bm{\omega}\right)$$ is the pseudo-gradient given by \eqref{eq:gradient_for_our}. We also define the singular value decomposition for $\M^{(k)}$ as $\M^{(k)} = \U^{(k)} \bm{D}^{(k)} \V^{(k)}$. 

There are three primary challenges. First, control the influence by using the pseudo-gradient $\widehat{\nabla} \mathcal{L}(\M;\Y,\bOmega)$ instead of the true gradient. Second, bound the impact of projecting $\N^{(k+1)}$ onto the rank $s$ matrix $\M^{(k+1)}$. Finally, maintain the incoherent condition throughout the iterative process. We examine the scenario where $s = r$ and establish bounds for the iterative series in the proposed algorithms. 

We begin our algorithm analysis with the selection of the initial value $\M^{(0)}$. Since $p^{-1} P_{\bOmega} (\Y)$ is an unbiased estimator of $\M_0$, a commonly used initial value for matrix completion problems is the rank-$r$ projection of this estimator, i.e., 
$$\M^{(0)} = \operatorname{Proj}_r(p^{-1} P_{\bOmega}(\Y)).$$ 
 Error bounds for the Frobenius and infinity norms, as well as an incoherent bound for $\M^{(0)}$, are provided in Lemma \ref{lem:bound_for_initial}.
\begin{lemma}
    \label{lem:bound_for_initial}
    With probability larger than $1-cn^{-3}$, it holds that
    \begin{align*}
        &\|\M^{(0)} - \M_0\|_{F} \le C(\rho^{-1/4}\|\M_0\|_{\infty} +\sigma)\kappa \sqrt{\frac{mr}{p}}\\
        &\|\M^{(0)} - \M_0\|_{\infty} \le \|\U\|_{2,\infty}\|\V\|_{2,\infty} (\|\M^{(0)} - \M_0\|_{F}+\sqrt{r}\|\bH\|)(1+f^{(0)})^2\\
        &\|\U^{(0)}\|_{2,\infty} \le \|\U\|_{2,\infty}(1+f^{(0)}),\quad \|\V^{(0)}\|_{2,\infty} \le \|\V\|_{2,\infty}(1+f^{(0)}),
    \end{align*}
    where
    \begin{align*}
        f^{(0)} =   C_1\sqrt{r}(\kappa - 1) + C_2(\kappa+\sqrt{r})(\|\M_0\|_{\infty}+6\sigma c\log m)^{1/2}\left(\frac{m}{p\sigma_r^2}\right)^{1/2} \log m.
    \end{align*}
\end{lemma}
In Lemma \ref{lem:bound_for_initial}, we demonstrate that $\M^{(0)}$ serves as a suitable initial matrix. It is sufficiently close to $\M_0$ and preserves the crucial incoherence condition without excessively increasing the incoherence coefficient. Using mathematical induction, we establish the main result below that tracks both the error $\M^{(k)} - \M_0$ and the incoherence condition for $\M^{(k)},\ k=1,2,\cdots$.
\begin{theorem}
    \label{thm:track_interation}
   With probability larger than $1-cn^{-3}$, for each $M^{(k)}$ in the iterative series $\left\{\M^{(k)}\right\}_{k=1,2,\cdots}$ with $\|\M^{(k)} - \M_0\| \ge C_3\kappa \sigma \sqrt{m/p}$, it holds that
   \begin{align*}
       & \|\M^{(k)} - \M_0\|_{F} \le \|\M^{(0)} - \M_0\|_{F}\prod_{i=0}^{k-1}\left(1 - \gamma_i(1-\alpha_i)\right),  \\
       & \|\U^{(k)}\|_{2,\infty}  \le  \|\U^{(0)}\|_{2,\infty}\prod_{i=0}^{k-1} \left(1 +C \mu_0 r^{4}\frac{\|\M^{(i)}-\M_0\|_F}{\sigma_r^{(k)}}\gamma_i\alpha_i\log^{3/2} m\right),\\
      &  \|\V^{(k)}\|_{2,\infty} \le   \|\V^{(0)}\|_{2,\infty}\prod_{i=0}^{k-1} \left(1 +C \mu_0 r^{4} \frac{\|\M^{(i)}-\M_0\|_F}{\sigma_r^{(k)}}\gamma_i\alpha_i\log^{3/2} m\right),\\
     & \|\M^{(k)} - \M_0\|_{\infty} \le 2\|\M^{(k)} - \M_0\|_{F}\|\U^{(k)}\|_{2,\infty}\|\V^{(k)}\|_{2,\infty},
   \end{align*}
   where we define that $\sigma_r^{(k)}$ is the $r$th singular value of $\M^{(k)}$, $\kappa_r^{(k)}$ is the condition number, $C^{(k)}_2<1$ is a sufficiently small constant and
    \begin{align*}
     \gamma_k &= \frac{2\eta_{k+1} }{\widehat{\sigma}(\M^{(k)})^2}\frac{(\|P_{\bOmega}(\M^{(k)}-\Y)\|- \widehat{\sigma}(\M^{(k)})\lambda_1)}{\|\M^{(k)} - \M_0\|_{F}},
   \\     \alpha_k &= \left(C^{(k)}_2 + \left(\frac{(\sigma^{(k)}_r\kappa^{(k)}_r)^{2}(2 - 2r(\|\M^{(k)} - \M_0\|_{F}/\sigma^{(k)}_r)^{-2})^2}{\|\M^{(k)} - \M_0\|_{F}^2} +1\right)^{-1/2}\right).
    \end{align*}
\end{theorem}

\begin{remark}
    When treating $\rho = 1-\gamma_i(1-\alpha_i) < 1$ as a constant, we obtain an exponentially decaying sequence such that $\|\M^{(k)} - \M_0\|_{F} \le \rho^k\|\M^{(0)} - \M_0\|_{F}$ holds if $\|\M^{(k)} - \M_0\| \ge C_3\kappa \sigma \sqrt{m/p}$. Theorem 2 in \cite{ma2018implicit} provides a similar decay for the gradient descent of the problem \eqref{eq:nonconvex-estimate}. A primary challenge it addresses is the ill-posed representation $\M = \bm{L}\bm{R}'$ which lacks identifiability.  In our theorem, we omit the matrix factorization and the alternating optimal step. We mainly focus on the issues related to the non-convex loss $\mathcal{L}(\M;\Y,\bOmega)$ in our proposed method. Expanding our results to the alternating gradient descent algorithm with the representation $\M = \bm{L}\bm{R}'$ is straightforward by integrating their techniques.
\end{remark}

In Theorem \ref{thm:track_interation}, we demonstrate that the Frobenius and infinity norms of $\M^{(k)} - \M_0$ decrease successively while controlling the growth rates of $\|\U^{(k)}\|_{2,\infty}$ and $\|\V^{(k)}\|_{2,\infty}$. This ensures that with a finite step $k$, we can obtain an $\M^{(k)}$ with small errors in both the Frobenius and infinity norms while maintaining the incoherence condition with a small incoherence coefficient.
\begin{corollary}
    \label{cor:iteration_results}
    If $p \gtrsim \mu_0^4\kappa^4r^{11} m^{-1}\log^{\alpha} m$, for sufficiently small step sizes $\eta_{k},k=1,2,\cdots$, there exist $K>0$ with probability larger than $1-cn^{-3}$ that
     \begin{align*}
       & \|\M^{(K)} - \M_0\|_{F} \lesssim \kappa \sigma\sqrt{mr/p},\\
        &\|\M^{(K)} - \M_0\|_{\infty} \lesssim  \kappa \sigma\mu_0 \sqrt{r^3\log^3 m/np}.
    \end{align*}
\end{corollary}
\begin{remark}
    Corollary \ref{cor:iteration_results} demonstrates that $\M^{(K)}$ is a reliable approximation for the estimator $\widetilde{\M}$. It has the same order in Frobenius error bounds as $\widetilde{\M}$, as established in Theorem \ref{thm:upper_bound_for_rankr}. The additional assumption in Corollary \ref{cor:iteration_results} is slightly more stringent than Assumption \ref{assum:missing}. Treating $\mu_0$, $\kappa$, and $r$ as constants, the lower bound of observation probability $p \gtrsim m^{-1}\log^{\alpha} m$ still aligns with the conditions specified in Assumption \ref{assum:missing}.
\end{remark}

Corollary \ref{cor:iteration_results} confirms that the algorithmic error does not dominate the statistical error. Our proposed algorithm achieves this by ensuring a sufficiently small step size in each update and an adequate number of iterations. As a result, there is no gap in the error bound rates between the oracle estimators and the results obtained through numerical methods. This is important because it guarantees the numerical performance of the proposed algorithm for our method.

\section{Numerical results}
\label{sec:numerical}
To show the numerical performance of our method, we compare the proposed estimators $\widetilde{\M}_{\text{fac}}$ and $\widetilde{\M}_{\text{nuc}}$, presented in equations \eqref{eq:our_estimator_nonconvex} and \eqref{eq:our_estimator_convex}, to the Frobenius norm minimizing estimators $\widehat{\M}_{\text{fac}}$ and $\widehat{\M}_{\text{nuc}}$, described in equations \eqref{eq:nonconvex-estimate} and \eqref{eq:convex-estimate}. This comparison is conducted using both simulation and real data experiments. For clarity, we denote $\widetilde{\M}_{\text{fac}}$ and $\widetilde{\M}_{\text{nuc}}$ as Estimator 1 and Estimator 2, with their corresponding baselines $\widehat{\M}_{\text{fac}}$ and $\widehat{\M}_{\text{nuc}}$ being Baseline 1 and Baseline 2. 

For the implementation details, we select the initial value $\M^{(0)} = \operatorname{Proj}_s(p^{-1} \Y)$ for $\widetilde{\M}_{\text{fac}}$ and $\widehat{\M}_{\text{fac}}$, which represents the rank-$s$ projection of the unbiased estimator $p^{-1} \Y$. We also set $\M^{(0)} = p^{-1} \Y$ as the initial value for $\widetilde{\M}_{\text{nuc}}$ and $\widehat{\M}_{\text{nuc}}$. Subsequently, we apply the alternating gradient descent algorithm (Algorithm 2 in \cite{ma2018implicit}) to obtain $\widehat{\M}_{\text{fac}}$ and use the singular value shrinkage method to obtain $\widehat{\M}_{\text{nuc}}$. Our estimators are calculated using the proposed algorithms, utilizing the pseudo-gradient \eqref{eq:gradient_for_our} in gradient descent step. Three types of error are considered to measure performance of the estimated $\M$: 1) Frobenius norm error, $\mathcal{E}_{F} = \|\M-\M_0\|_F/\sqrt{mn}$; 2) Spectral norm error, $\mathcal{E}_{sp} = \|\M-\M_0\|/\sqrt{m}$; and 3) Maximum norm error, $\mathcal{E}_{\infty} = \|\M-\M_0\|_{\infty}$. 

\subsection{Simulation study}
In simulations, the observation matrix $\Y\in \mathbb{R}^{m\times n}$ is generated by $\Y = P_{\bOmega}(\M_0 +\bH)$. Here $\bOmega$ is an index matrix with entries following i.i.d. binomial distributions with probability $p$.
Each entry in the noise matrix $\bH$ is independently generated as the sum of a mean zero Gaussian random variable with variance $\sigma^2/2$ and a Rademacher random variable multiplied by $\sigma/\sqrt{2}$. The underlying true matrix $\M_0$ is generated using the singular value decomposition $\M_0 = \U_0 \bm{D}_0 \V_0'$. Here $\U_0\in \mathbb{R}^{m\times r}$ and $\V_0\in \mathbb{R}^{n\times r}$ are the leading $r$ left and right singular vectors of a randomly generated standard Gaussian matrix $\bm{W}\in\mathbb{R}^{m\times n}$. The delocalization property of random Gaussian matrices ensures that $\M_0$ is $\mu$-incoherent with a small $\mu$. Here $\bm{D}_0\in \mathbb{R}^{r\times r}$ is a diagonal matrix with $[\bm{D}_0]_{i,i} = \sigma_i,1\le i \le r$, the singular values of $\M_0$. Three different true ranks of $\M_0$ are chosen: $r = 5, 10, 20$, and the singular values $\sigma_i,1\le i\le r$ are set as
\begin{align*}
    \text{Case 1: }&r = 5,\ \ 
     \quad \sigma_i = (3 +i/3)  \sqrt{m}(1+\sqrt{n/m}),\quad 1\le i \le 5;\\
    \text{Case 2: }&r = 10,\quad \sigma_i = (3 +i/3)  \sqrt{m}(1+\sqrt{n/m}),\quad 1\le i \le 10;\\
    \text{Case 3: }&r = 20,\quad \sigma_i = (3 +(i+1)/6)  \sqrt{m}(1+\sqrt{n/m}),\quad 1\le i \le 20.
\end{align*} 

\begin{figure}[t]
  \vspace{-0.25cm}
  \centering 
  \subfigbottomskip=2pt 
  \subfigcapskip=-50em
  \subfigure{
  \includegraphics[width=0.32\linewidth]{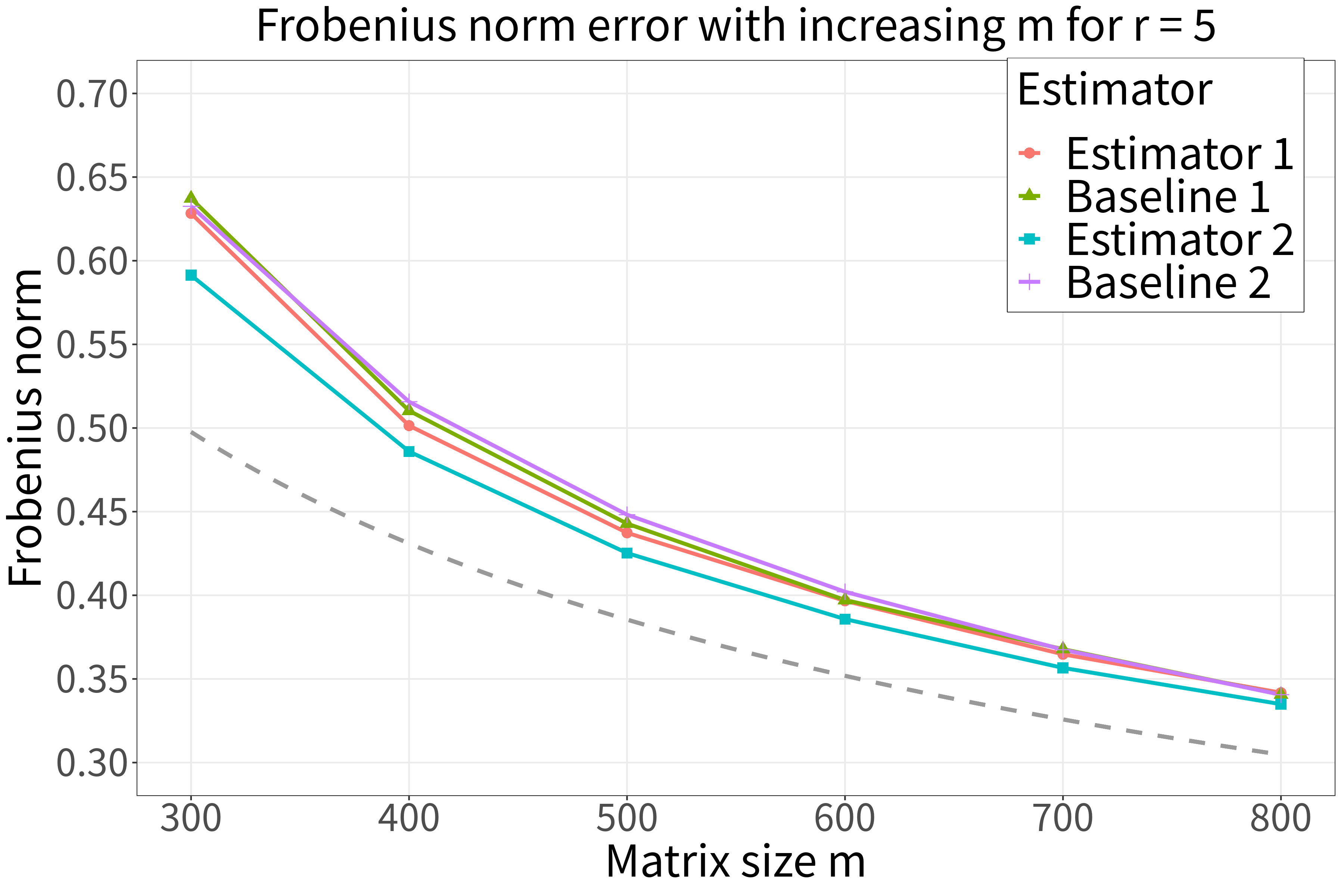}}
  \subfigure{
  \includegraphics[width=0.32\linewidth]{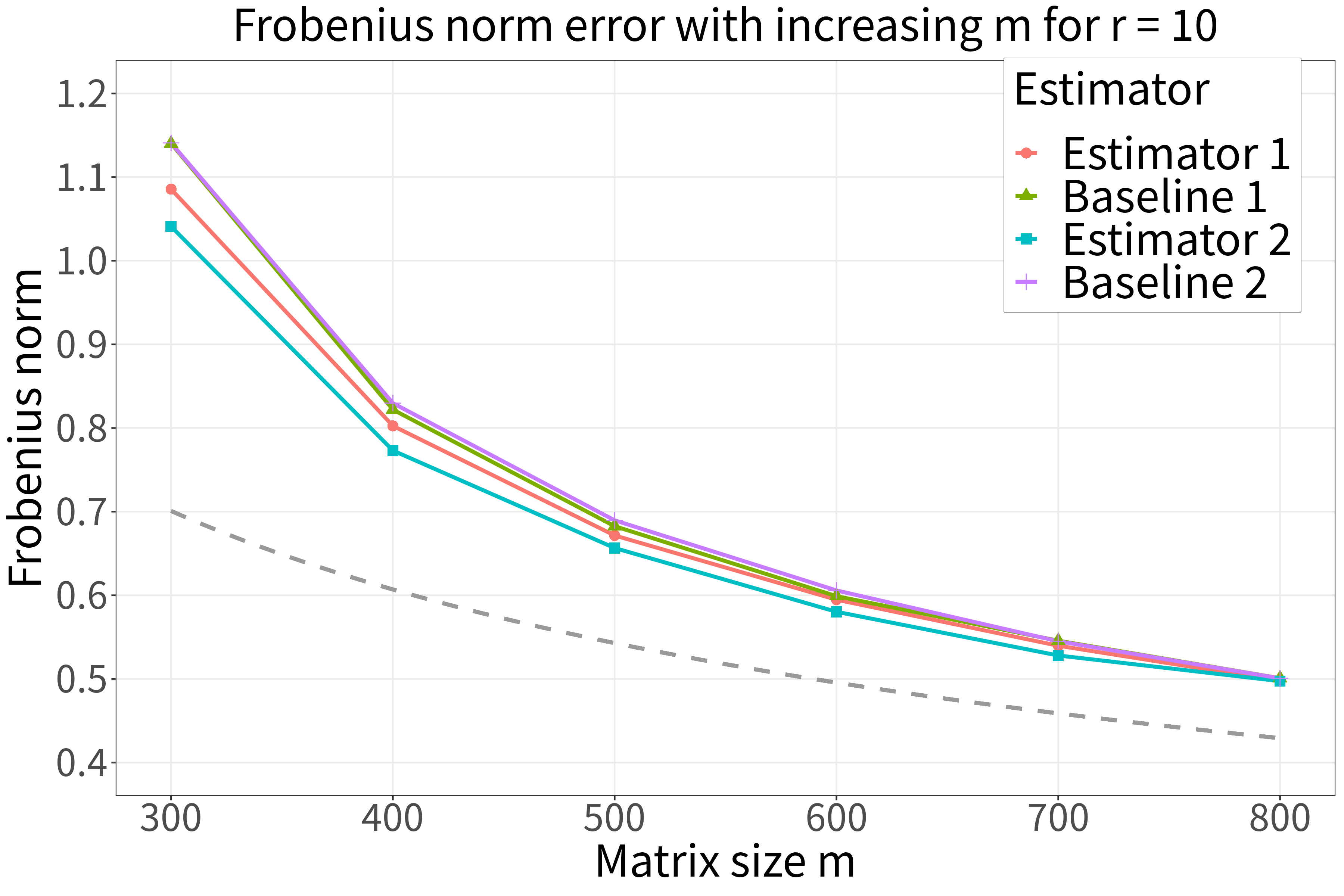}}
  \subfigure{
  \includegraphics[width=0.32\linewidth]{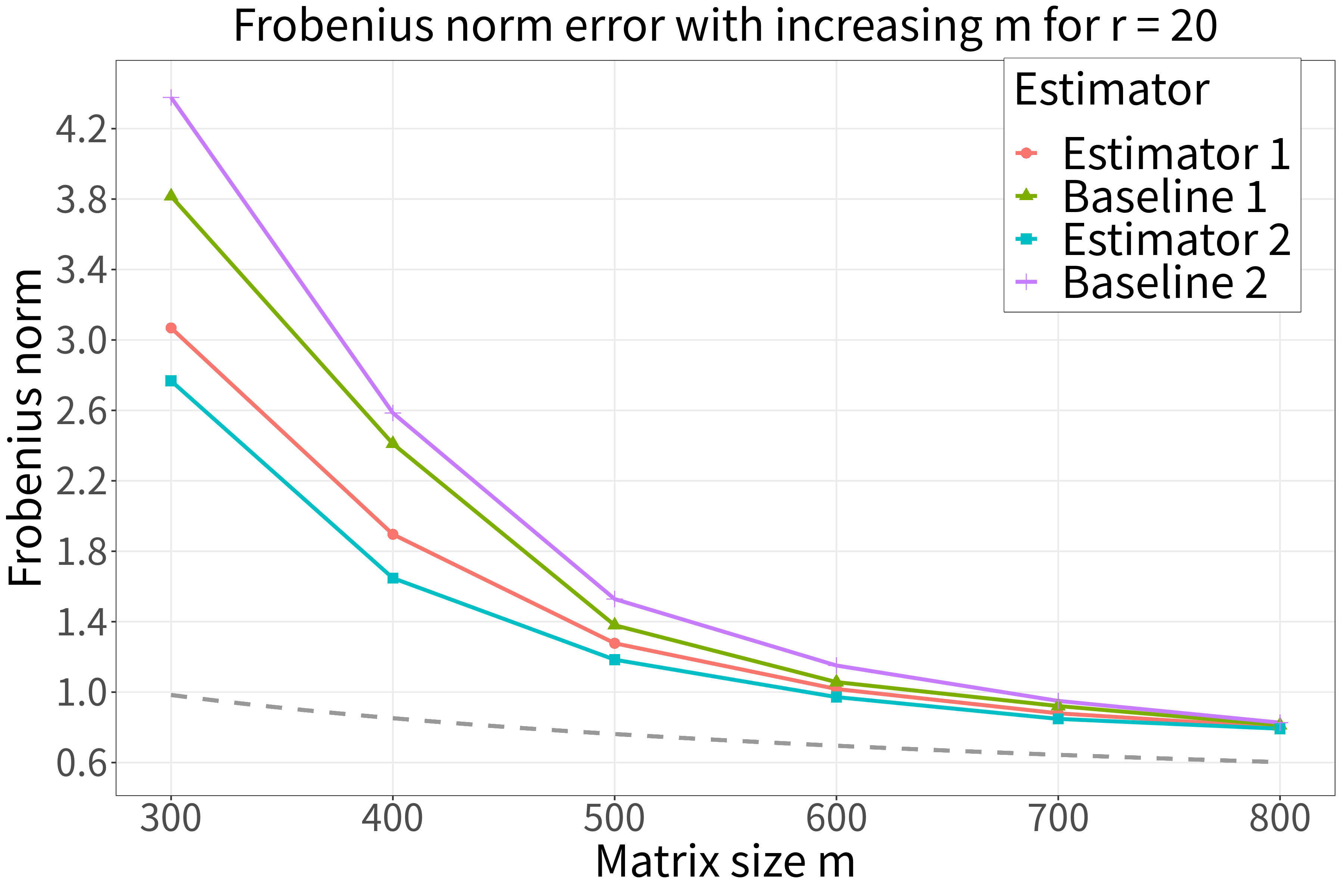}}
 \caption{The trend curve of the Frobenius norm error $\mathcal{E}_F$ for our estimators and baselines when matrix size approaches infinity. The plots from left to right correspond to the scenarios $r=5,10,20$.}
  \label{fig:change_matrix_size}
  \vspace{-0.25cm}
  \end{figure}

Firstly, we examine how the matrix sizes $m,n$ of $\M_0$ affect errors of the proposed estimators. We set the standard deviation of $\bH$ to $\sigma = 1$ and the observation probability $p$ to $0.2$. For the estimators $\widetilde{\M}_{\text{fac}}$ and $\widehat{\M}_{\text{fac}}$, we use $s = r$. The observed data is randomly divided into $80\%$ for training and $20\%$ for validation. The tuning parameter $\lambda$ in calculating $\widetilde{\M}_{\text{nuc}}$ and $\widehat{\M}_{\text{nuc}}$ is determined by minimizing the mean squared error on the validation set. In Table \ref{tab:change_matrix_size}, we present the average Frobenius norm error $\mathcal{E}_{F}$ from $100$ repeated runs, varying the matrix size $m$ from $200$ to $800$ with $n/m=0.5$. The proposed estimators $\widetilde{\M}_{\text{fac}}$ and $\widetilde{\M}_{\text{nuc}}$ demonstrate lower Frobenius norm error compared to corresponding baselines. The average spectral norm error $\mathcal{E}_{sp}$ and maximum norm error $\mathcal{E}_{\infty}$ show similar results and are provided in the supplementary materials due to space limit. In Figure \ref{fig:change_matrix_size}, we illustrate the trend of the Frobenius norm error as the matrix size $m$ approaches infinity. The dashed line represents $p^{-1/2}$ times the Frobenius norm error for $\widecheck{\M}_{\text{rmt}}$ in the matrix denoising model, serving as the optimal lower bound of order $\sigma\sqrt{r/mp}$. As $m$ increases, errors of the proposed methods decrease following the theoretical order, which coincide with the theoretical findings  in Theorem \ref{thm:upper_bound_for_rankr} and Corollary \ref{cor:convex_tuning_lambda}. 

\begin{table}[t]
 \vspace{-0.25cm}
    \caption{The Frobenius norm error $\mathcal{E}_{F}$ of proposed methods and corresponding baselines for $\sigma=1$, observation probability $p=0.2$, rank of $\M_0$ as $5,10,20$ and matrix size $m$ from $200$ to $800$ with $n/m=0.5$.}
    \label{tab:change_matrix_size}
    \begin{tabular}{c c |c c c c c c c}
		\toprule[1pt]
		 Rank &Estimator & m=200&m=300&m=400&m=500&m=600&m=700&m=800\\
    \midrule[1pt]
     \multirow{4}{*}{r = 5}&Estimator 1: $\widetilde{\M}_{\text{fac}}$&  0.8422  &0.6283&0.5014&0.4374&0.3967&0.3647&0.3418\\ 
     &Baseline 1: $\widehat{\M}_{\text{fac}}$&0.9007 &0.6372&0.5103&0.4429&0.3972&0.3678&0.3406\\ 
     \cline{2-9}
      &Estimator 2: $\widetilde{\M}_{\text{nuc}}$& 0.8108 &0.5913&0.4860&0.4252&0.3858&0.3566&0.3349\\ 
       &Baseline 2: $\widehat{\M}_{\text{nuc}}$& 0.9052&0.6325&0.5157&0.4482&0.4021&0.3675&0.3406\\ 
    \hline
         \multirow{4}{*}{r = 10}&Estimator 1: $\widetilde{\M}_{\text{fac}}$&  2.5518 & 1.0856 & 0.8025 &0.6715 &0.5947 &0.5397 &0.4977 \\ 
     &Baseline 1: $\widehat{\M}_{\text{fac}}$& 3.1489 & 1.1402&0.8220 &0.6826 & 0.5989& 0.5457& 0.5008\\ 
     \cline{2-9}
      &Estimator 2: $\widetilde{\M}_{\text{nuc}}$&  2.1028 & 1.0409&0.7731& 0.6565& 0.5803&0.5280 &0.4974 \\ 
       &Baseline 2: $\widehat{\M}_{\text{nuc}}$&2.7178  & 1.1408&0.8297 &0.6896 &0.6058 &0.5450 &0.5007 \\ 
    \hline
         \multirow{4}{*}{r = 20}&Estimator 1: $\widetilde{\M}_{\text{fac}}$& 3.7134 & 3.0684 & 1.8963&1.2775 & 1.0178& 0.8797& 0.7999\\ 
     &Baseline 1: $\widehat{\M}_{\text{fac}}$& 3.8373 &3.8161 &2.4105 & 1.3799&1.0568 & 0.9210&0.8125 \\ 
     \cline{2-9}
      &Estimator 2: $\widetilde{\M}_{\text{nuc}}$&  3.7677&2.7677 &1.6476 &1.1842 &0.9719 & 0.8480& 0.7921\\ 
       &Baseline 2: $\widehat{\M}_{\text{nuc}}$& 4.4844 & 4.3773& 2.5853& 1.5286& 1.1511& 0.9500&0.8267 \\ 
    \bottomrule[1pt]
    	\end{tabular}
     \vspace{-0.25cm}
     \end{table}

\begin{figure}[t]
  \vspace{-0.25cm}
  \centering 
  \subfigbottomskip=2pt 
  \subfigcapskip=-50em
  \subfigure{
  \includegraphics[width=0.4\linewidth]{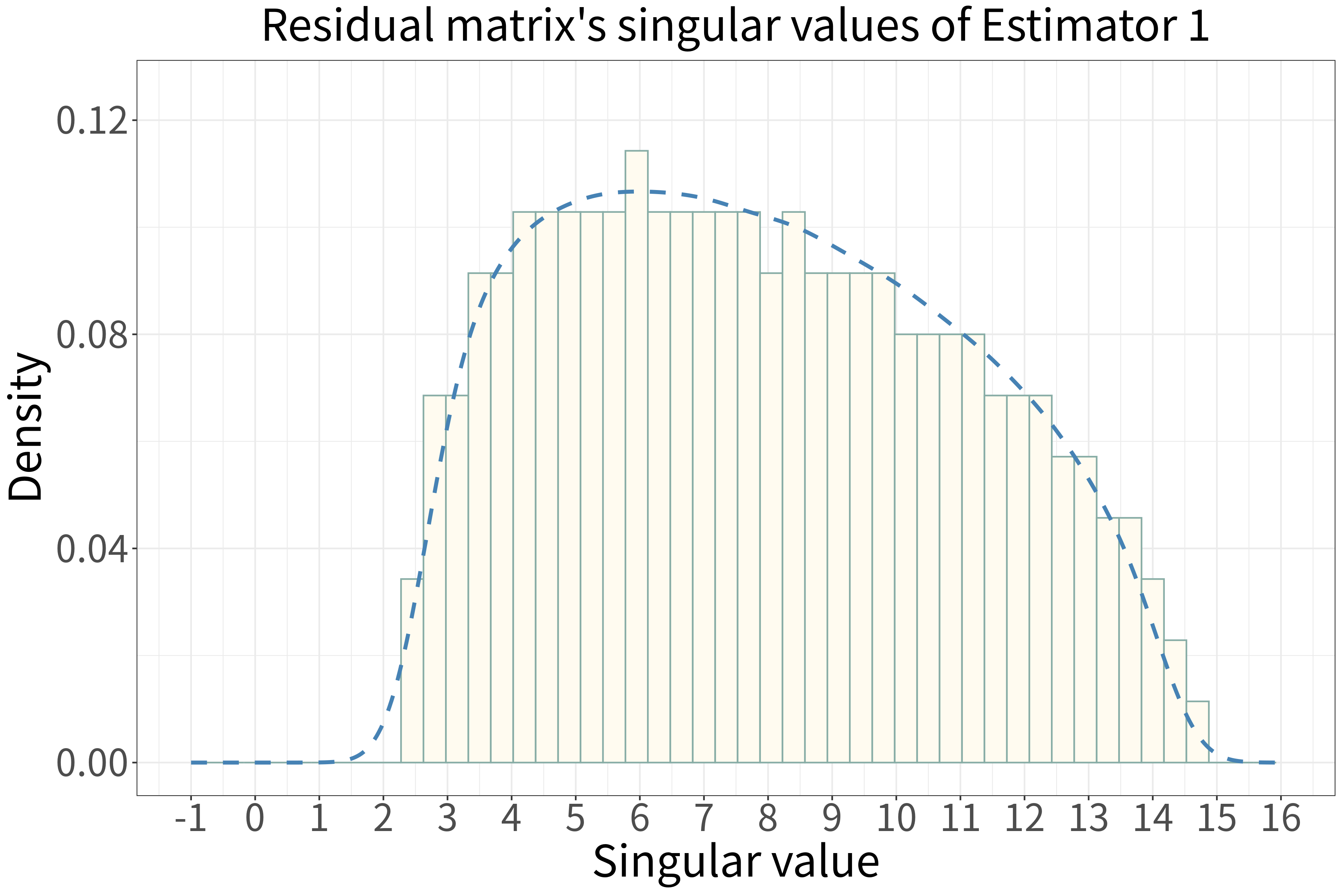}}
  \subfigure{
  \includegraphics[width=0.4\linewidth]{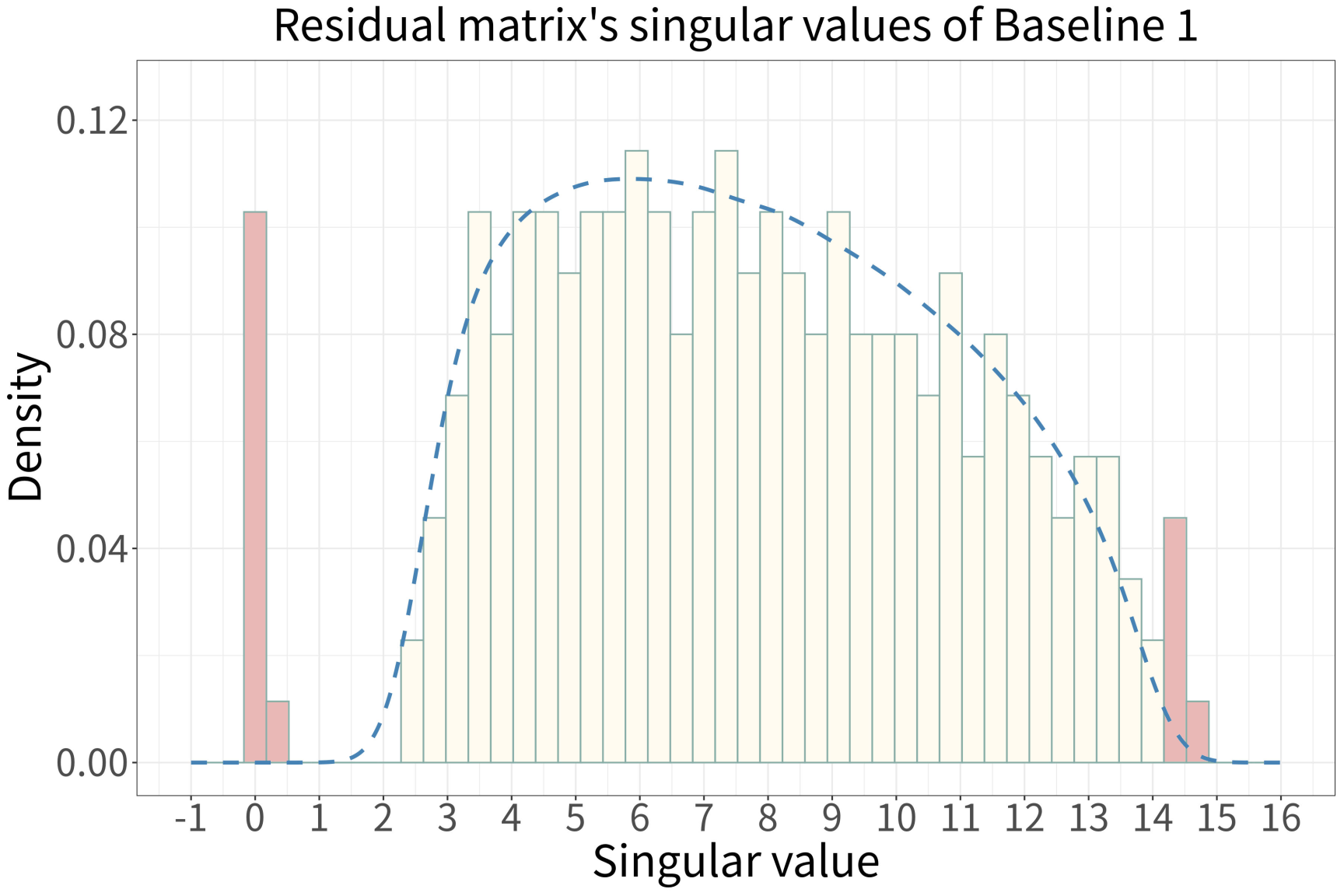}}
  
  \subfigure{
  \includegraphics[width=0.4\linewidth]{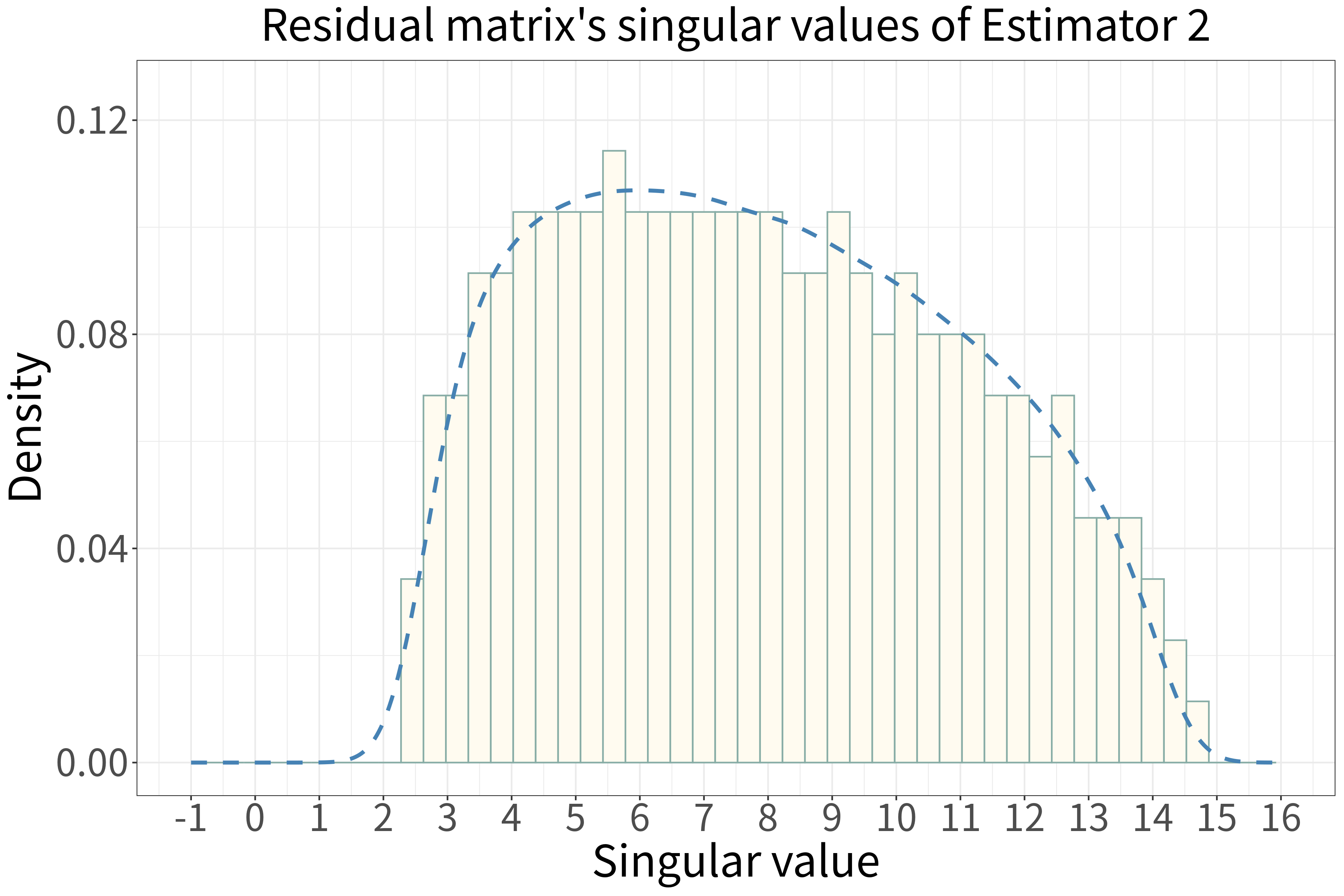}}
    \subfigure{
  \includegraphics[width=0.4\linewidth]{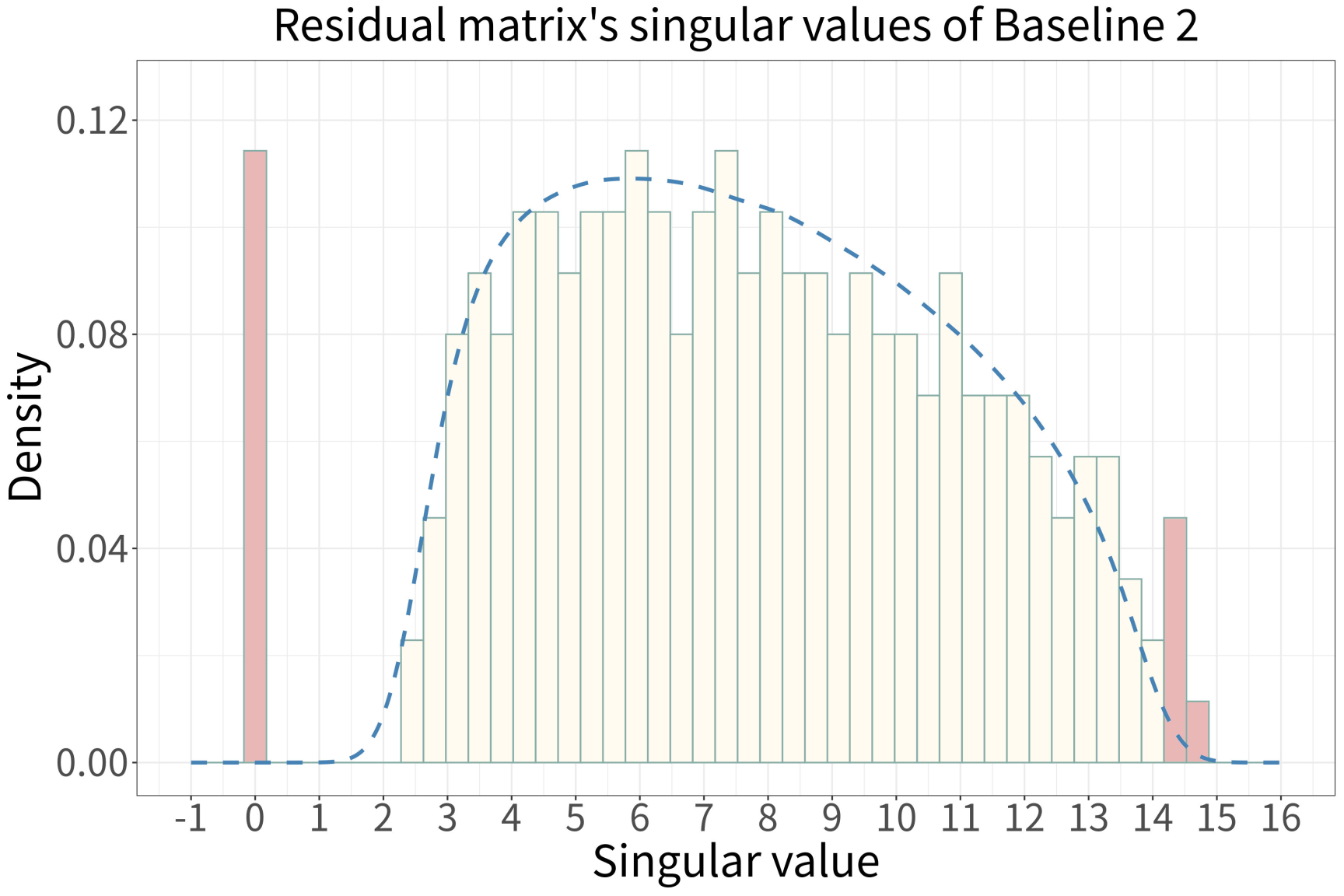}}
 \caption{The singular value distributions of residual matrix for our estimators and baselines in the scenarios $r = 10$ and matrix size $500\times 250$. The left two plots correspond to the results of Estimator 1 and Baseline 1, and the right two plots correspond to the results of Estimator 2 and Baseline 2.}
  \label{fig:residual_matrix}
  \vspace{-0.25cm}
  \end{figure}

Secondly, we examine the singular value distributions of residual matrices in both the proposed and baseline methods. Figure \ref{fig:residual_matrix} illustrates that the empirical spectral distributions of the residual matrices for our estimators closely resemble the expected spectral distribution of sparse random matrices, scaled by the estimated working standard deviation \eqref{eq:working_variance}. In contrast, the residual matrices in the baselines deviate from the behavior of a sparse random matrix, exhibiting outliers in both the top and bottom singular values.

\begin{figure}[t]
  \vspace{-0.25cm}
  \centering 
  \subfigbottomskip=2pt 
  \subfigcapskip=-50em
  \subfigure{
  \includegraphics[width=0.32\linewidth]{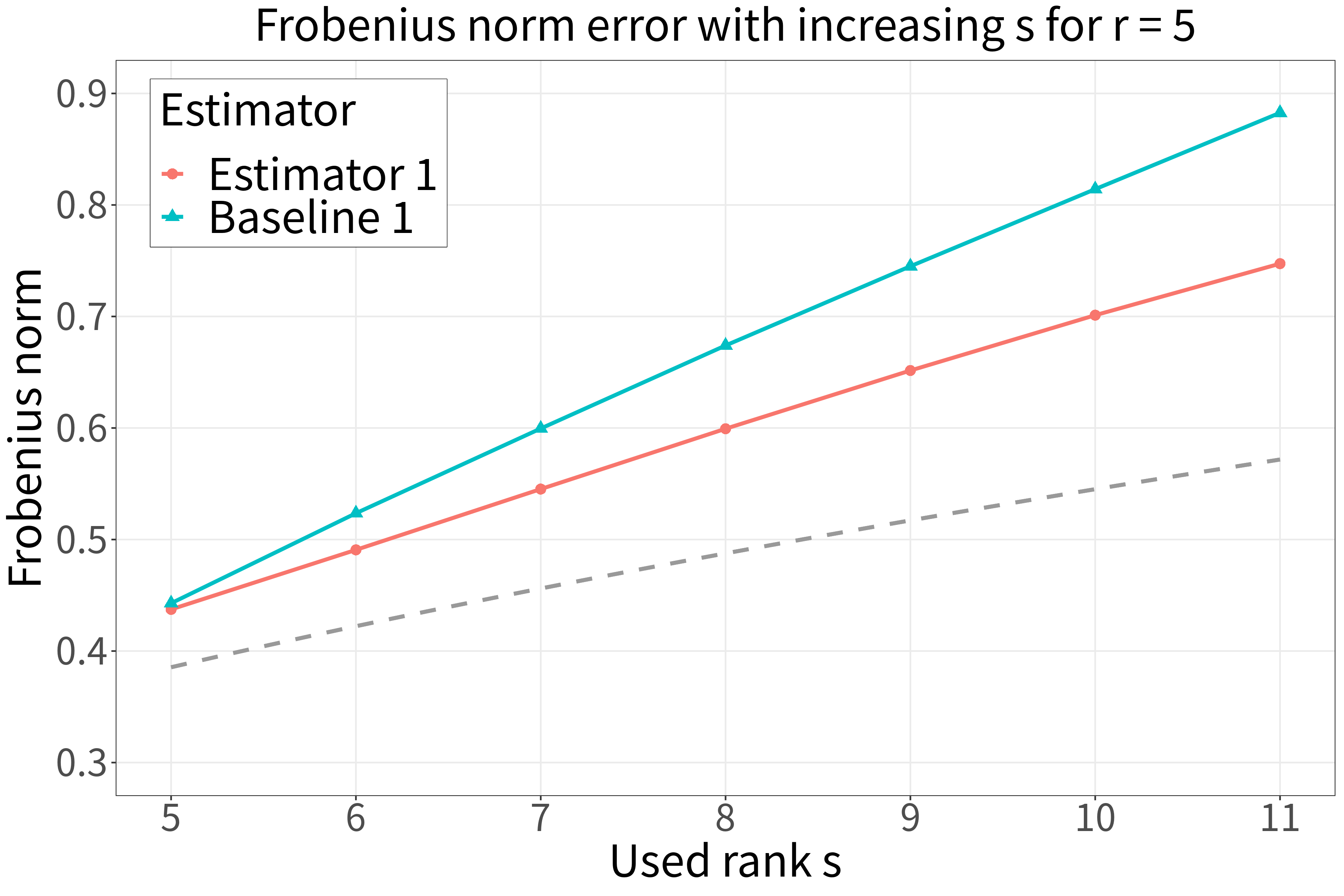}}
  \subfigure{
  \includegraphics[width=0.32\linewidth]{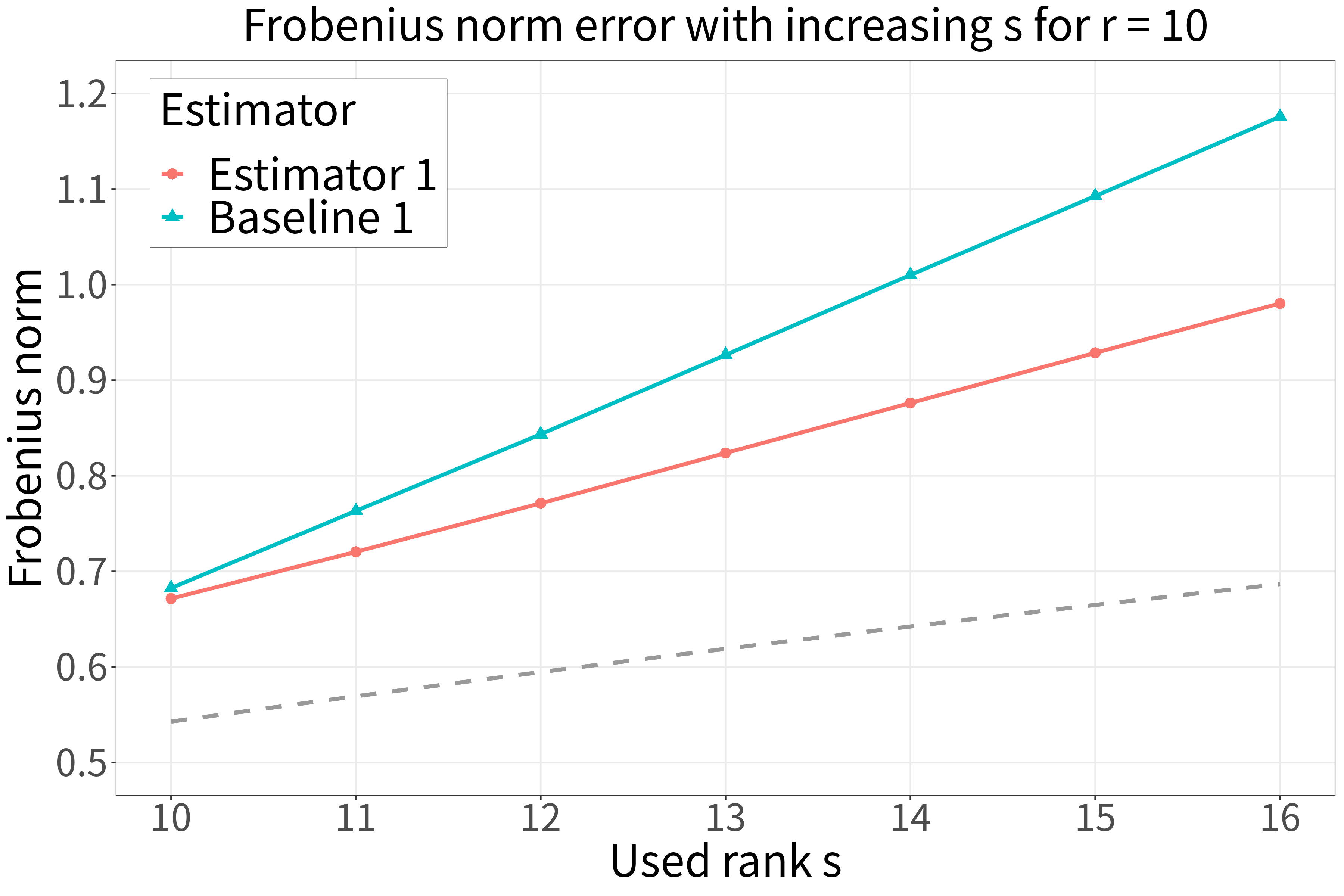}}
  \subfigure{
  \includegraphics[width=0.32\linewidth]{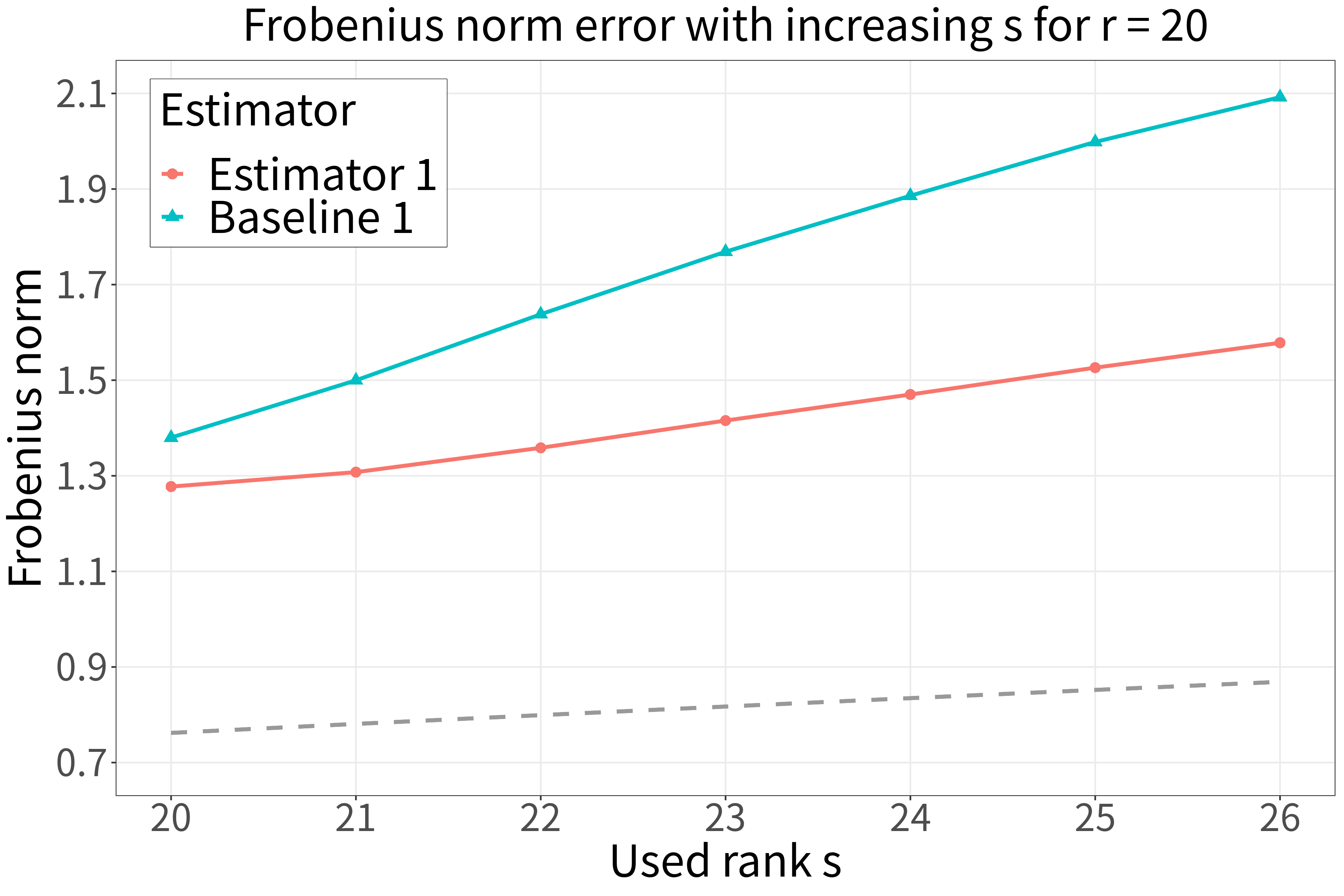}}
 \caption{The trend curve of the Frobenius norm error $\mathcal{E}_F$ for Estimator 1 and Baseline 1 when $s\ge r$ increases. The plots from left to right correspond to the scenarios $r=5,10,20$.}
  \label{fig:change_s}
  \vspace{-0.25cm}
  \end{figure}

Then, we investigate the influence of misspecifying the true rank in Estimator 1 and Baseline 1, i.e., the effect of the parameter $s$ on the estimators $\widetilde{\M}_{\text{fac}}$ and $\widehat{\M}_{\text{fac}}$. Table \ref{tab:change_s} presents the average Frobenius norm error $\mathcal{E}_F$ for $100$ repeated runs, with $s$ varying from the true rank $r$ to $r+6$, while keeping all other settings as previously described. As $s$ gradually deviates from the true rank $r$, the estimation errors of both Estimator 1 and Baseline 1 increase. However, the error of our estimator rises at a significantly slower rate, demonstrating greater robustness, which is evident shown by Figure \ref{fig:change_s}. The dashed line represents the behavior of $\widecheck{\M}_{\text{rmt}}$, similar to what is shown in Figure \ref{fig:change_matrix_size}. The average spectral norm error $\mathcal{E}_{sp}$ and maximum norm error $\mathcal{E}_{\infty}$ show similar patterns and are also provided in the supplementary materials. It shows that as $s\ge r$ increases, the Frobenius norm error remains approximately at the theoretical order of $\sigma\sqrt{(s+r)/mp}$, consistent with the findings in Corollary \ref{cor:mis_ranks}. 
\begin{table}[t]
\vspace{-0.25cm}
    \caption{The Frobenius norm error $\mathcal{E}_{F}$ of Estimator 1 and Baseline 1 for true rank of $\M_0$ as $5,10,20$, used rank $s$ from $r$ to $r +6$ and matrix size $500\times250$.}
    \label{tab:change_s}
    \begin{tabular}{c c |c c c c c c c}
		\toprule[1pt]
		 Rank &Estimator& s = r&s = r+1& s = r+2&s = r+3&s = r+4&s = r+5&s = r+6\\
    \midrule[1pt]
     \multirow{2}{*}{r = 5}&Estimator 1: $\widetilde{\M}_{\text{fac}}$&  0.4374&0.4907&0.5452&0.5993&0.6516&0.7012&0.7474\\ 
     &Baseline 1: $\widehat{\M}_{\text{fac}}$&0.4429&0.5236&0.5996&0.6740&0.7451&0.8142&0.8827\\
    \hline
         \multirow{2}{*}{r = 10}&Estimator 1: $\widetilde{\M}_{\text{fac}}$& 0.6715  & 0.7205 & 0.7712 &0.8238 & 0.8761 & 0.9286 & 0.9804\\ 
     &Baseline 1: $\widehat{\M}_{\text{fac}}$& 0.6826 & 0.7633&0.8435 &0.9265 &1.0101 & 1.0927& 1.1757\\ 
    \hline
         \multirow{2}{*}{r = 20}&Estimator 1: $\widetilde{\M}_{\text{fac}}$& 1.2775 & 1.3076 &1.3585 & 1.4156& 1.4701& 1.5262&1.5784 \\ 
     &Baseline 1: $\widehat{\M}_{\text{fac}}$& 1.3799 & 1.4997 & 1.6380&1.7688 &1.8860 &1.9985 &2.0922 \\ 
    \bottomrule[1pt]
    	\end{tabular}
\vspace{-0.25cm}
\end{table}

\begin{figure}[t]
  \vspace{-0.25cm}
  \centering 
  \subfigbottomskip=2pt 
  \subfigcapskip=-50em
  \subfigure{
  \includegraphics[width=0.32\linewidth]{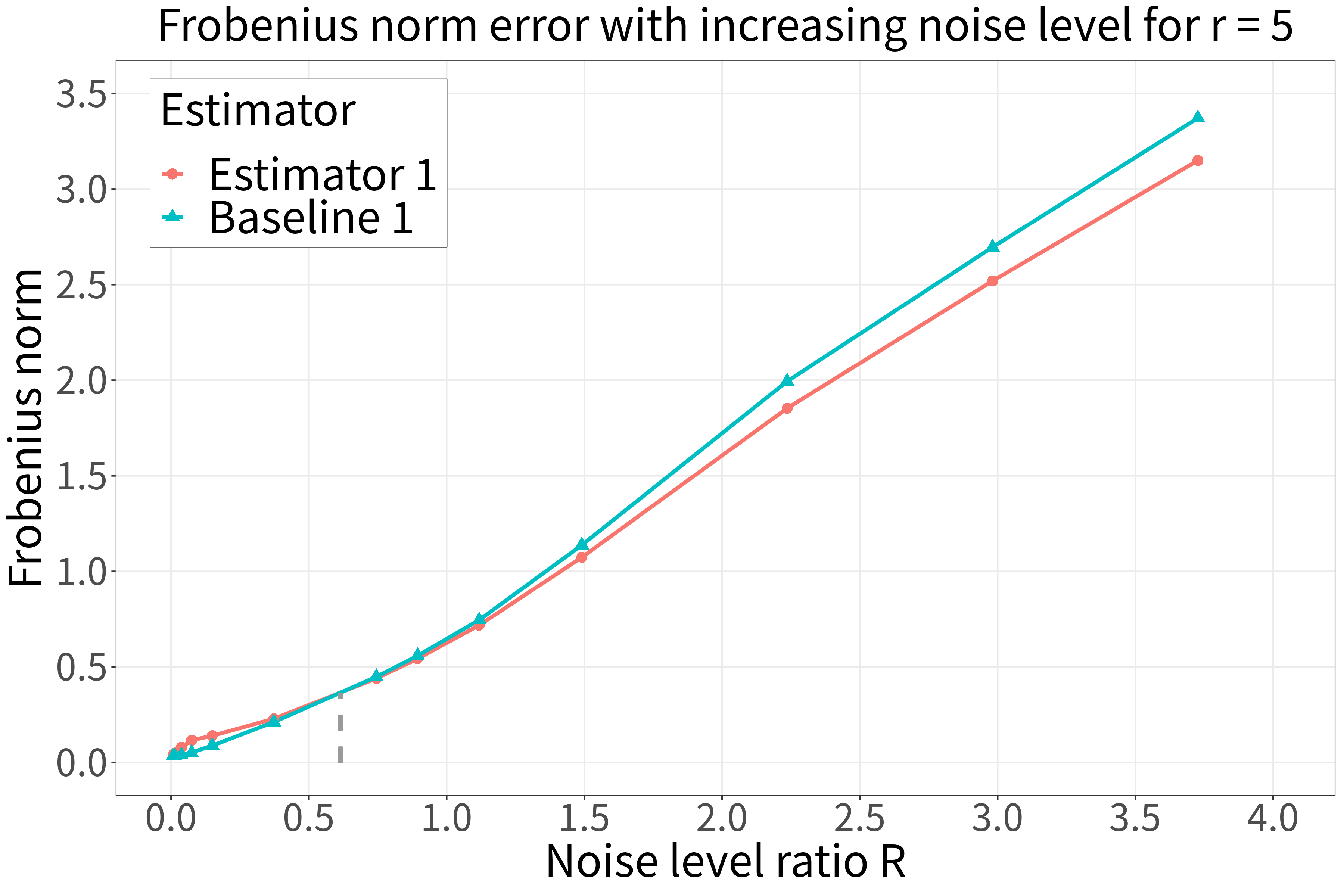}}
  \subfigure{
  \includegraphics[width=0.32\linewidth]{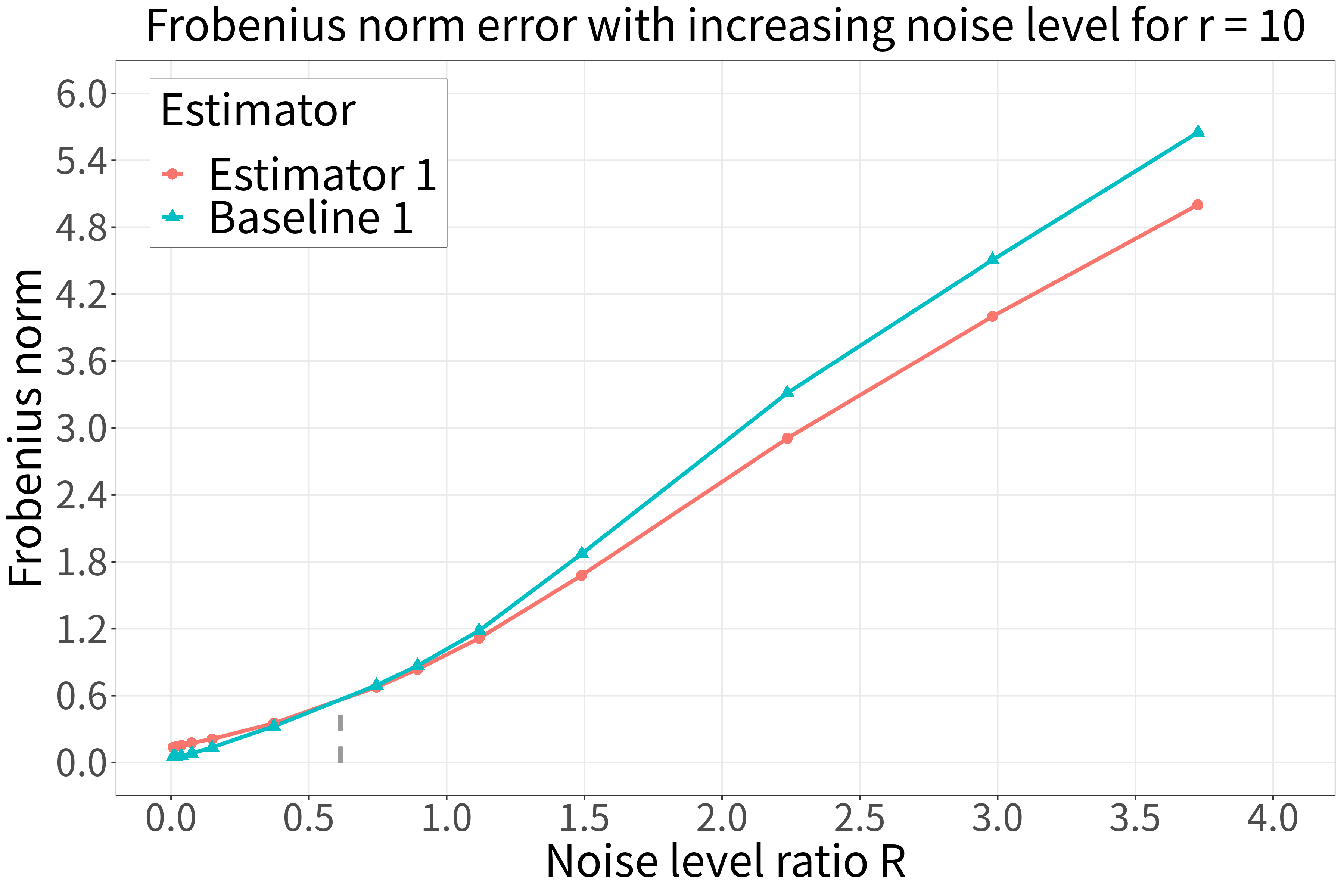}}
  \subfigure{
  \includegraphics[width=0.32\linewidth]{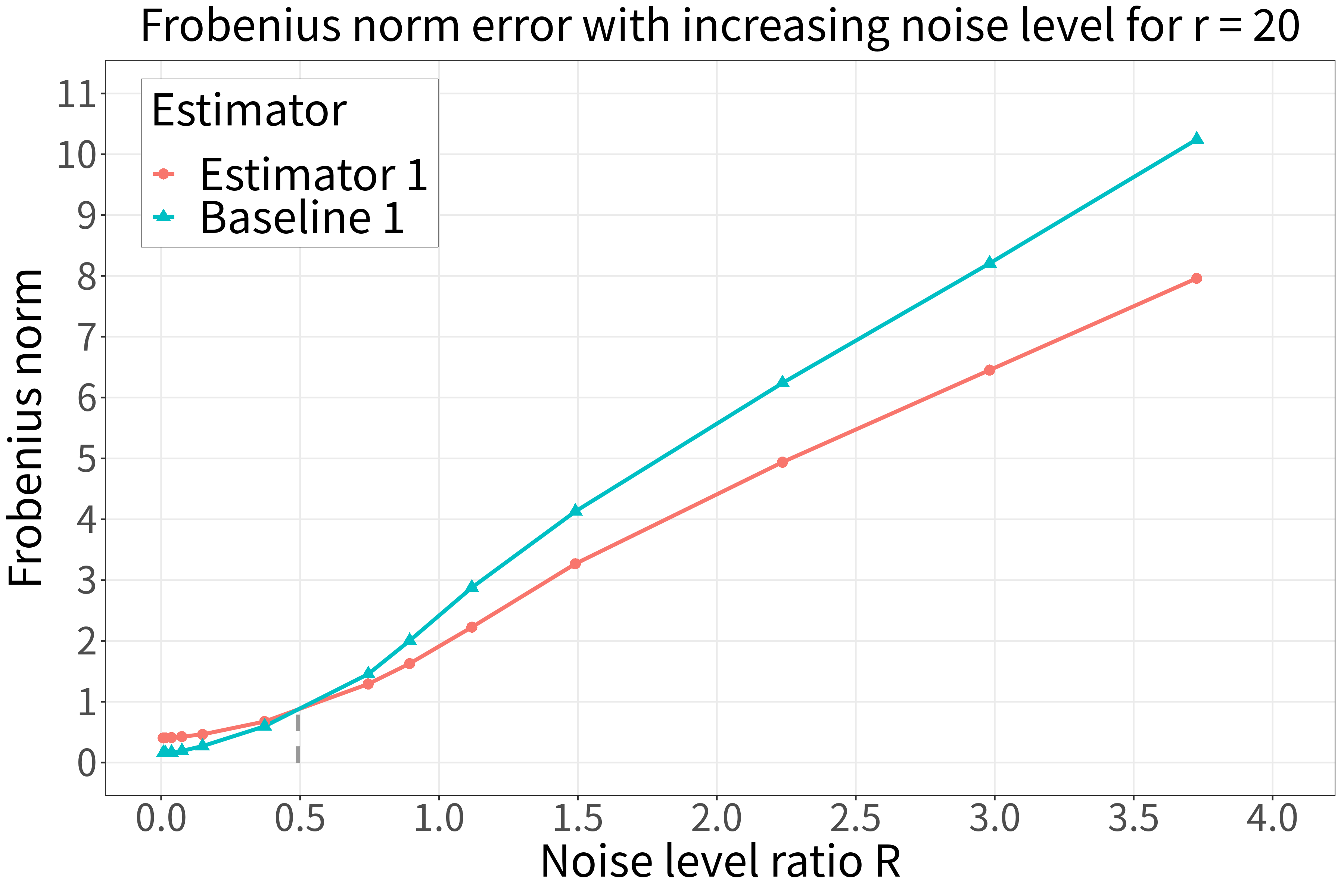}}
 \caption{The Frobenius norm error $\mathcal{E}_F$ for Estimator 1 and Baseline 1 when noise level ratio $R$ increases. The plots from left to right correspond to the scenarios $r=5,10,20$.}
  \label{fig:change_noise}
  \vspace{-0.25cm}
  \end{figure}

 Finally, we evaluate the effectiveness of our method compared to baselines under varying noise levels. We use the ratio of the the spectral norm of $\bH$, to the $r$th singular value of $\M_0$ multiplied by the rooted observation probability, as a metric for assessing the intensity of noise, i.e., $R =\|\bH\|/(p^{1/2}\sigma_r(\M_0))$. We fix $\M_0$ and vary the standard deviation $\sigma$ of the noise, from $0.01$ to $5$. In Figure \ref{fig:change_noise}, when the noise level is low (approximately $R \le 0.5$), the spectral information of the noise matrix $\bH$ is difficult to use, making Frobenius norm based methods slightly better. However, when the noise level increases ($R > 0.5$), our method performs significantly more effective. This indicates that our proposed method is especially suitable for scenarios with relatively high noise levels. 
\begin{figure}[t]
  \vspace{-0.25cm}
  \centering 
  \subfigbottomskip=2pt 
  \subfigcapskip=-50em
  \subfigure{
  \includegraphics[width=0.4\linewidth]{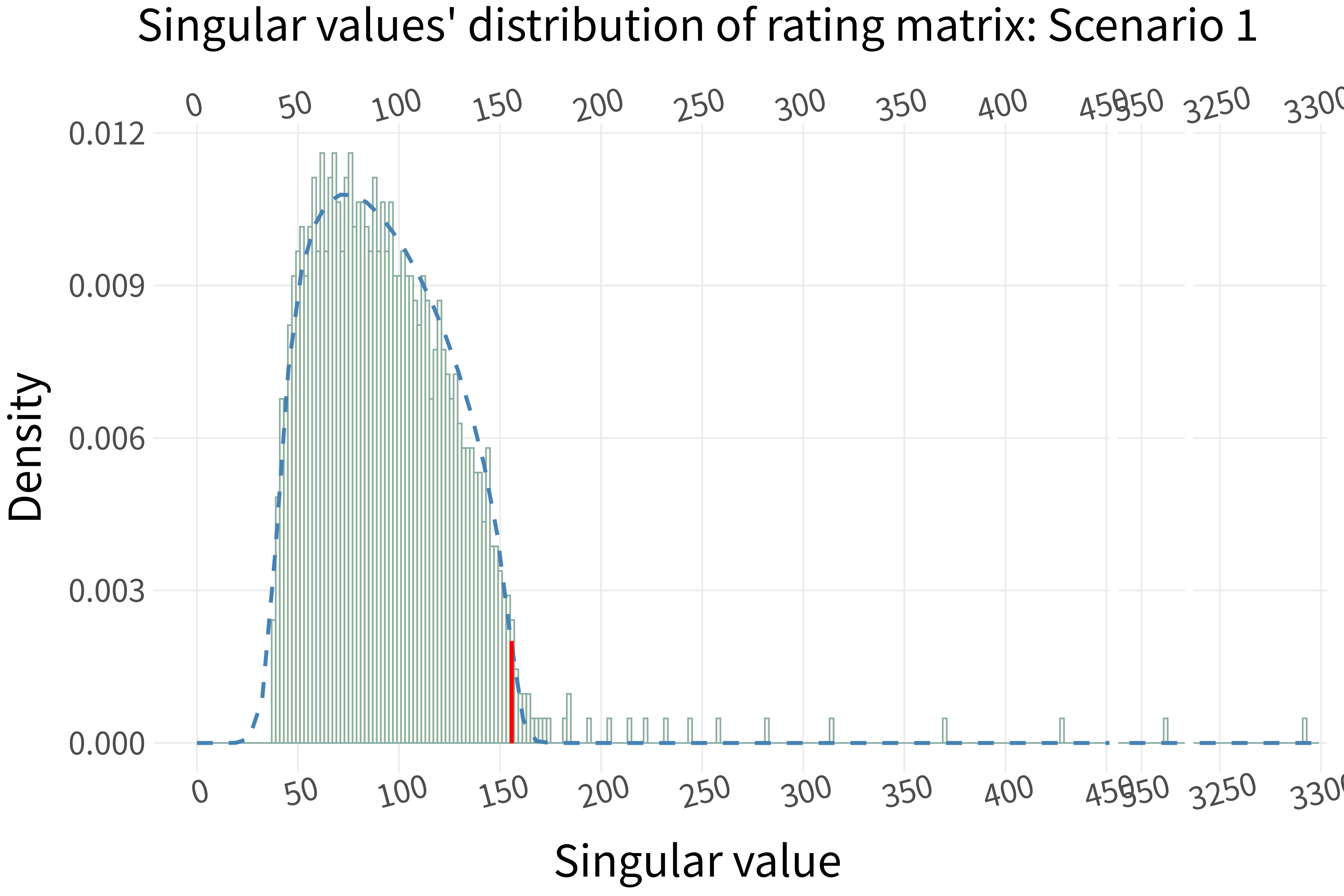}}
    \subfigure{
  \includegraphics[width=0.4\linewidth]{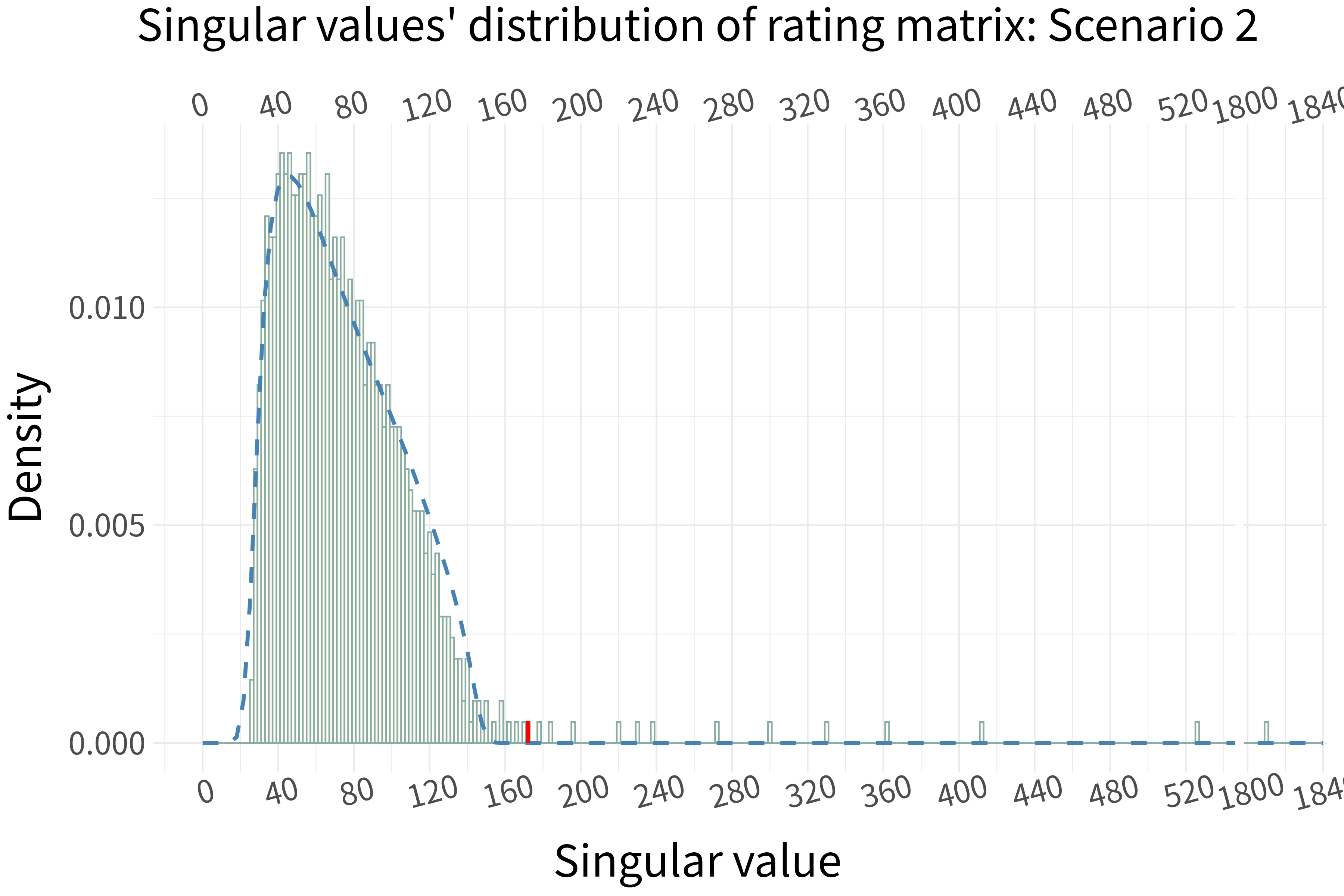}}
  
    \subfigure{
  \includegraphics[width=0.4\linewidth]{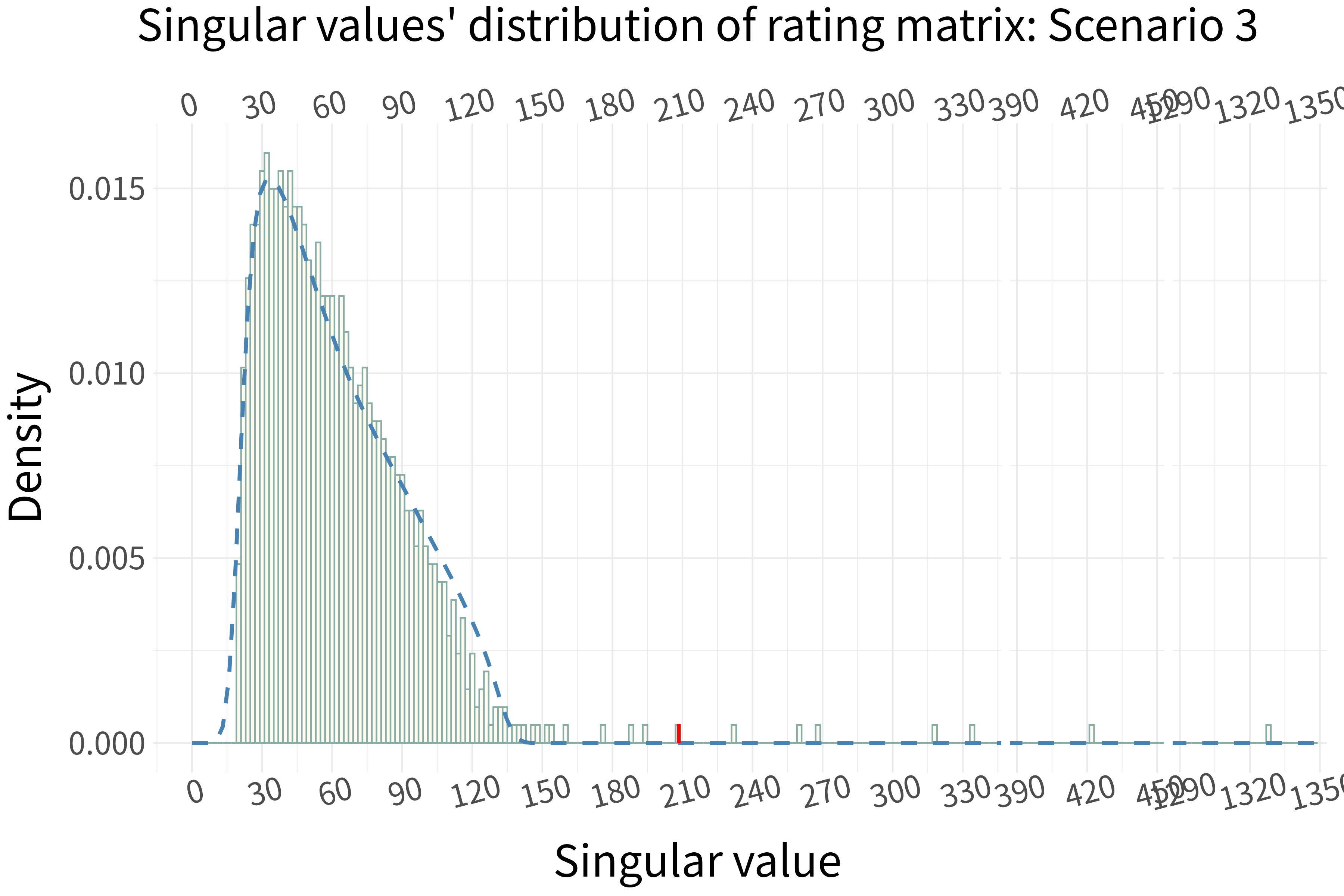}}
    \subfigure{
  \includegraphics[width=0.4\linewidth]{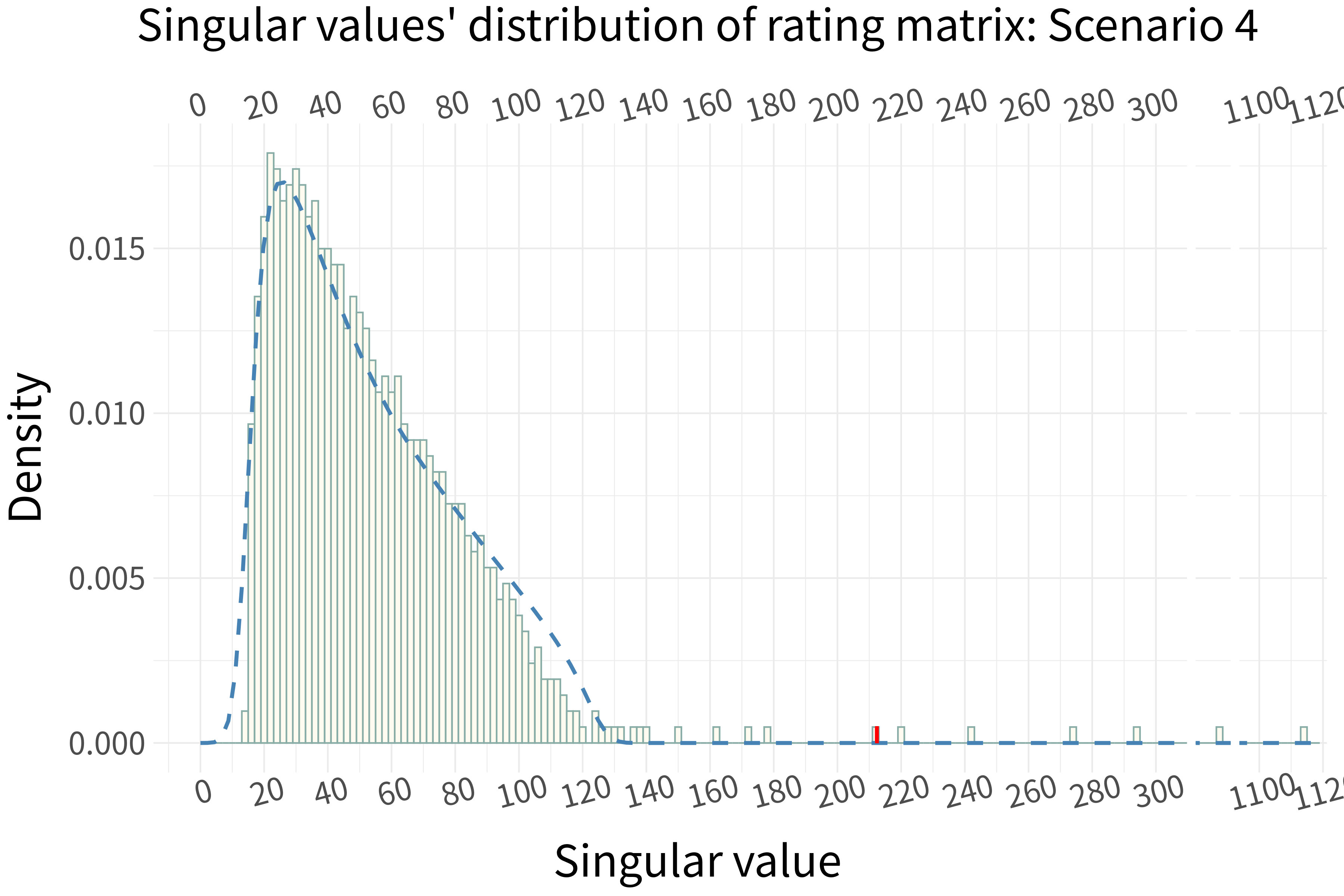}}
 \caption{The spectral distribution of the observed rating matrix and selected rank $s$ in each scenario respectively. The dash lines are densities of expectation of singular values of corresponding variance-adjusted sparse random matrices. The red lines shows the selected rank for Estimator 1.}
  \label{fig:spectral_distribution}
  \vspace{-0.25cm}
  \end{figure}
\subsection{Netflix rating data}
We utilize the Netflix Prize Dataset to demonstrate the performance of the proposed method. This dataset is detailed in \cite{bennett2007netflix} and is publicly available at \href{https://www.kaggle.com/datasets/netflix-inc/netflix-prize-data}{https://www.kaggle.com/datasets/netflix-inc/netflix-prize-data}. The dataset contains user ratings (ranging from 1 to 5) for various movies on Netflix from October 1998 to December 2005. Due to the vast number of movies available, each user has rated only a small portion of them. Applying matrix completion directly to this dataset allows for predicting users' ratings for movies they have not viewed, thereby developing a recommendation system. The dataset is preprocessed by retaining movies rated more than 25,000 times, excluding lesser-known films. Four distinct user groups are considered. In scenario 1, we focus on users who are actively engaged in watching and rating movies, retaining a total of 3,036 users, each with more than 700 ratings. In scenario 2 to 4, we consider general users and randomly sample 3,000 users from those with more than 200, 75 and 30 ratings, respectively. This results in a rating matrix with 3,036 users and 1,034 movies for scenario 1 and matrices with 3,000 users and 1,034 movies for scenarios 2 to 4. The observation probabilities in these rating matrices are approximately 0.50, 0.36, 0.23, and 0.12 for scenarios 1 to 4. Figure \ref{fig:spectral_distribution} shows the singular value distributions of the observed rating matrices for each scenario. In all cases, the distribution of singular values aligns closely with the expected distribution of the corresponding sparse random matrices after variance adjustment, with only a few outliers due to low-rank perturbations.

\begin{table}[t]
\vspace{-0.25cm}
    \caption{The RMSE in the test dataset for the proposed estimators, their corresponding baselines, and modified versions, evaluated across four scenarios of Netflix rating data.}
    \label{tab:netflix_data}
    \begin{tabular}{c c c c c |c c c c}
		\toprule[1pt]
		& Estim1& Mod-Estim1& Base1 & Mod-Base1 &Estim2  & Mod-Estim2  & Base2 &  Mod-Base2 \\
    \midrule[1pt]
    Scenario1 & 0.7680&  \textbf{0.7665}& 0.7775 & 0.7761 &0.7713 &  0.7688  &0.7722 & 0.7696 \\
    Scenario2 &0.8075 & \textbf{0.8058} & 0.8118 &0.8099 & 0.8119& 0.8100& 0.8148 &0.8123 \\
    Scenario3 &0.8383&\textbf{0.8368}  & 0.8421 & 0.8407 & 0.8467& 0.8443& 0.8504&0.8479 \\
    Scenario4 & 0.8490& \textbf{0.8468}&  0.8505 &0.8493 & 0.8670  & 0.8621  & 0.9053 & 0.8879 \\
    \bottomrule[1pt]
    	\end{tabular}
\vspace{-0.25cm}
\end{table}

\begin{figure}[t]
  \vspace{-0.25cm}
  \centering 
  \subfigbottomskip=2pt 
  \subfigcapskip=-50em
  \subfigure{
  \includegraphics[width=0.4\linewidth]{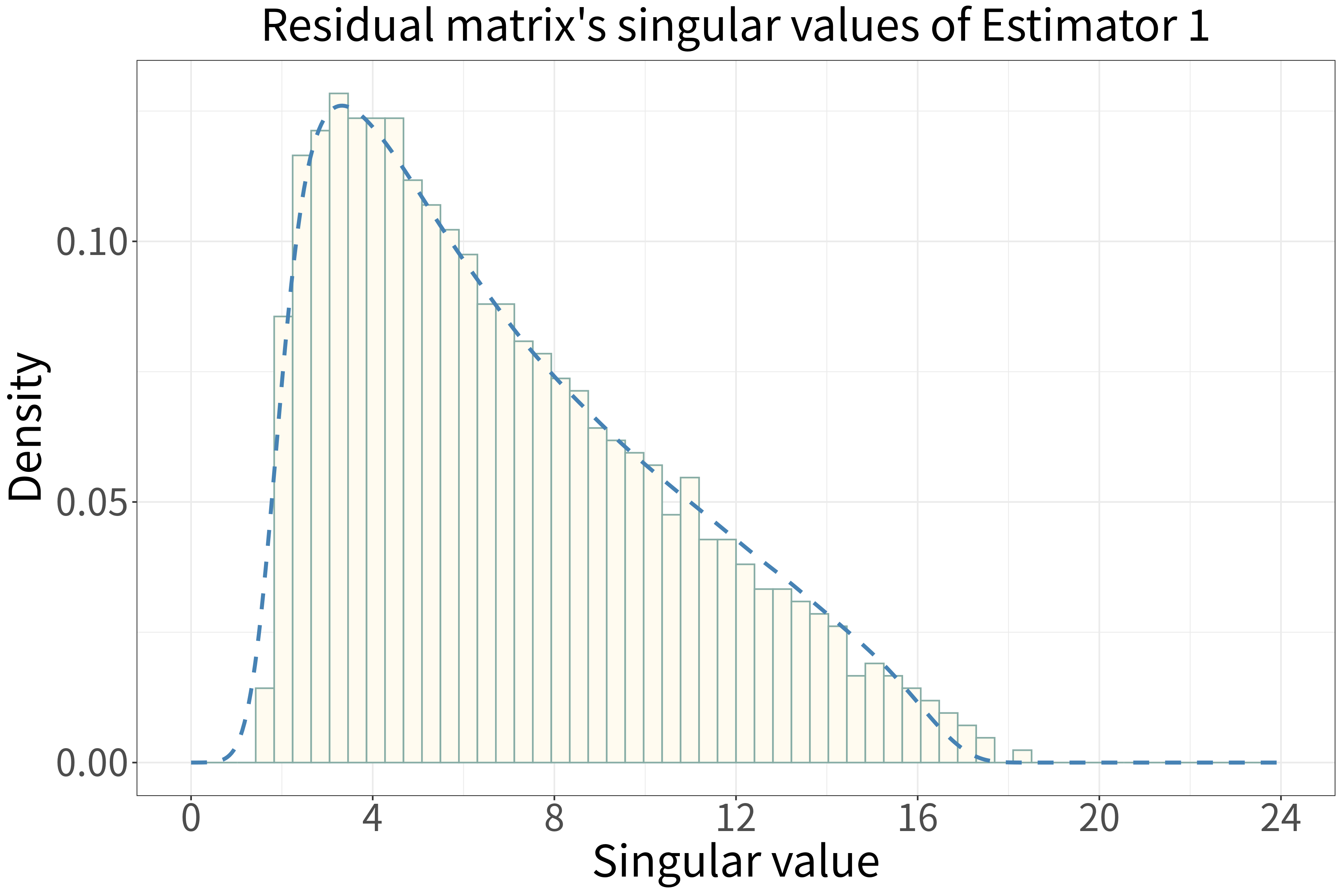}}
    \subfigure{
  \includegraphics[width=0.4\linewidth]{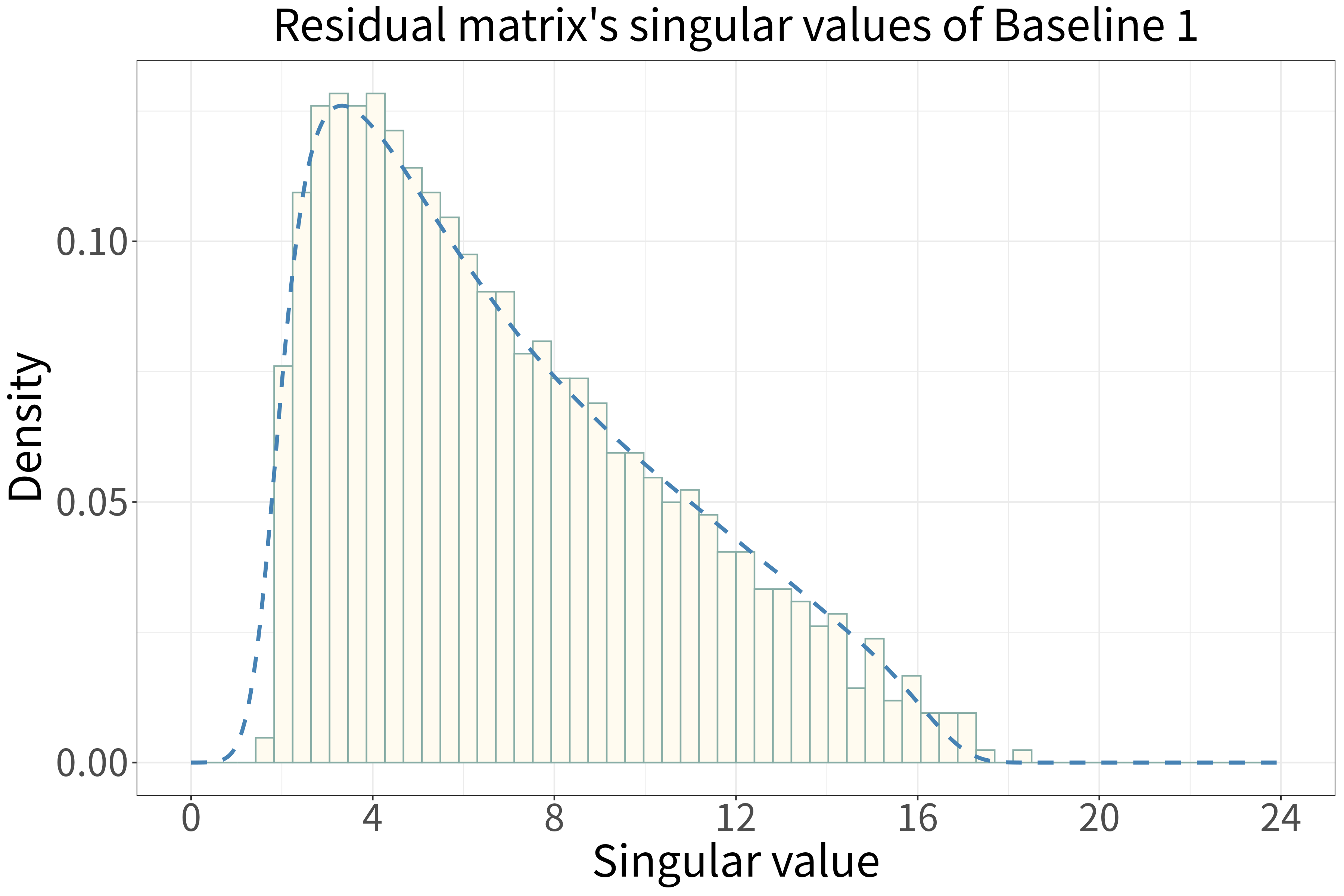}}
  
    \subfigure{
  \includegraphics[width=0.4\linewidth]{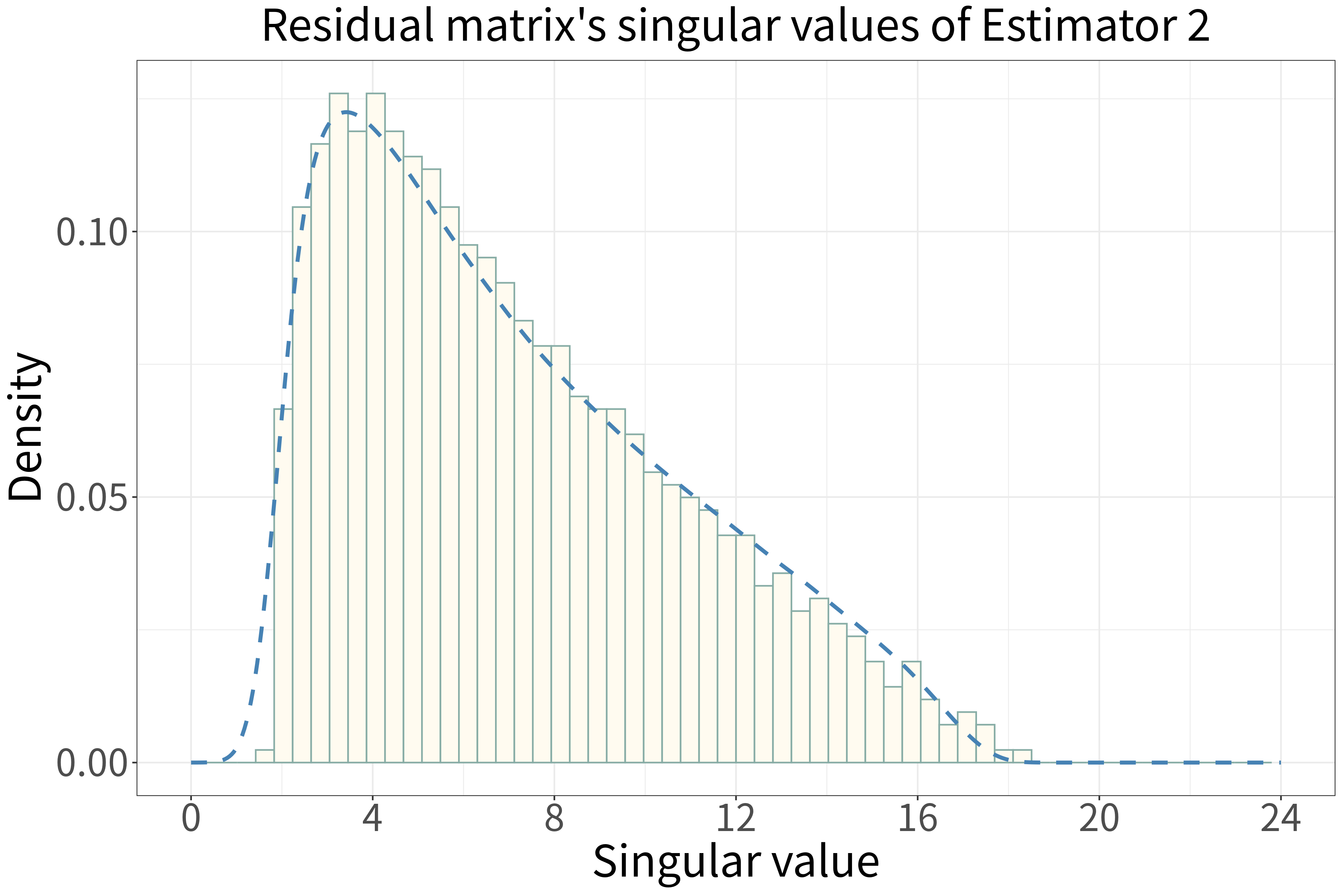}}
    \subfigure{
  \includegraphics[width=0.4\linewidth]{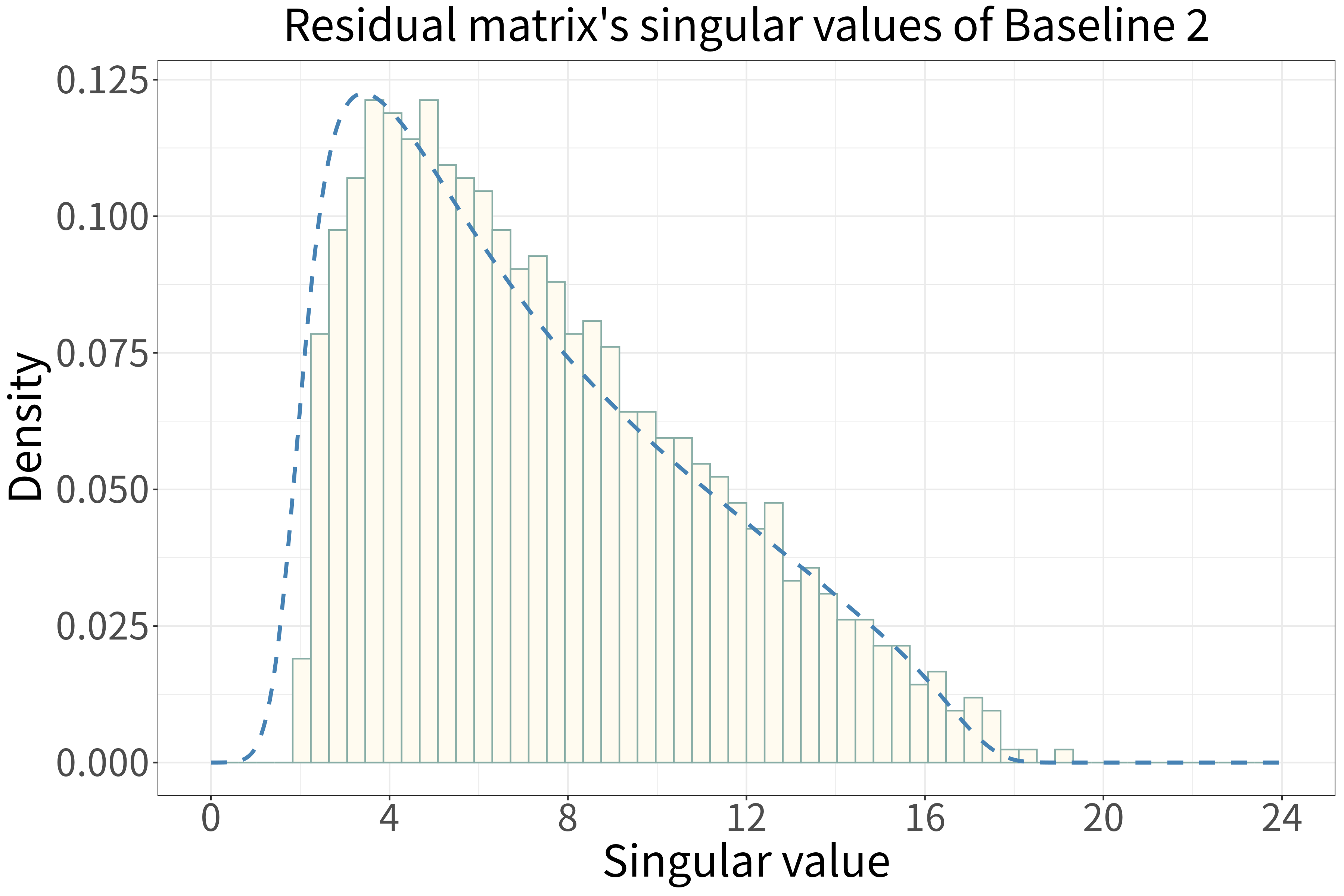}}
 \caption{The distribution of singular values of residual matrix for proposed and baseline methods for scenario 4. The left two plots correspond to the results of Estimator 1 and Baseline 1, and
the right two plots correspond to the results of Estimator 2 and Baseline 2. }
  \label{fig:spectral_distribution_residual}
  \vspace{-0.25cm}
  \end{figure}
The proposed and baseline methods are implemented in the four scenarios as described earlier. The dataset is divided into three parts: training, validation, and testing, with a ratio of $4:1:1$. Hyperparameters $s$ in Estimator 1 and Baseline 1, and $\lambda$ in Estimator 2 and Baseline 2, are tuned by minimizing the rooted mean squared error (RMSE) on the validation set. For scenarios 1 to 4, Estimator 1 selects ranks $s = 34, \ 14, \ 8, \ 7$ and Baseline 1 selects ranks $s = 34,\ 15, \ 8, \ 7$.  
Given that ratings fall within the range $[1,5]$, we truncate the estimated matrices to this interval, referring to them as modified versions. In scenarios 1 to 4, the noise level ratio $R$ can be approximately estimated by our estimators and the selected ranks, to be $R = 0.58, 0.50 ,0.48$ and $0.60$. This indicates that the Netflix data is subject to a relatively high noise level. Under high noise conditions ($R> 0.5$), our estimators consistently outperforms the baseline methods in the simulation study. For the Netflix rating data, Table \ref{tab:netflix_data} demonstrates that in all scenarios, the proposed estimators and their modified (Mod) versions yield smaller prediction errors on the test set, compared to the corresponding baselines. The empirical spectral distribution of the residual matrices from our estimators also closely aligns with the expected spectral distributions of the corresponding variance-adjusted sparse random matrices, showing a better fit than the baselines. Figure \ref{fig:spectral_distribution_residual} displays the singular value distributions of the residual matrices for both our estimators and the baselines in scenario 4 as an example.

  \subsection{Amazon reviews data} We also compare performance of our estimators with the baselines using the Amazon reviews 2023 dataset \citep{hou2024bridging}, which is publicly available at \href{https://amazon-reviews-2023.github.io}{https://amazon-reviews-2023.github.io}. The data collected user reviews in 2023 with rich features including the users rating, text and purchase verification. We consider four scenarios for reviews, each corresponding to a distinct category: movies, books, electronics and automotive. For each scenario, the top $1000$ items in each category are considered, with $3000$ users randomly selected after data preprocessing. Hence, Four $1000\times 3000$ reviewing matrices are observed with observation probabilities $0.0128,0.0127,0.0110, 0.0068$ for scenarios 1 to 4 respectively. The data preprocessing and implementation are similar to those for the Netflix rating data and we present the implementation details in the supplementary material. The noise level ratio $R$ for the four scenarios is around $R=1.16, 1.09, 1.54$ and $1.56$, which means that the dataset is under a higher noise level. Table \ref{tab:amazon_data} presents the RMSE on the test data for our estimators, their modified versions and the baselines. The performance gaps between our estimators (including modified versions) and the baselines are larger across all four scenarios.

\begin{table}[t]
\vspace{-0.25cm}
    \caption{The RMSE in the test dataset for the proposed estimators, their corresponding baselines, and modified versions, evaluated across four scenarios of Amazon review data.}
    \label{tab:amazon_data}
    \begin{tabular}{c c c c c |c c c c}
		\toprule[1pt]
		& Estim1& Mod-Estim1& Base1 & Mod-Base1 &Estim2  & Mod-Estim2  & Base2 &  Mod-Base2 \\
    \midrule[1pt]
    Scenario1 & 0.9264& 0.9176 & 0.9364 & 0.9272 & 0.9251 &  \textbf{0.9168} & 0.9457 & 0.9367 \\
    Scenario2 & 0.8441 & \textbf{0.8338} & 0.8686 & 0.8576 & 0.9239 &  0.8883 &0.8656 &0.8549  \\
    Scenario3 &0.9912 & \textbf{0.9805} & 1.0020 &0.9907  & 0.9912& 0.9810 &1.0180 &1.0072 \\
    Scenario4 & 0.9787 & \textbf{0.9588} &0.9898  & 0.9689 & 0.9831 & 0.9642  & 1.0172 &  0.9997 \\
    \bottomrule[1pt]
    	\end{tabular}
\vspace{-0.25cm}
\end{table}
\section{Discussion}
\label{sec:discussion}

This paper introduces a novel spectral matching criterion for noisy matrix completion. We demonstrate that estimators based on this criterion perform well both theoretically and numerically. Efficient algorithms are devised for the proposed criterion, ensuring small Frobenius and maximum norms. Our methods provide advantages over commonly used least squared estimators by utilizing more information from noise, especially in high noise level situations.

Our method can be readily generalized to encompass broader missing mechanisms and more complex noise matrices if the variance structure of the noise is understood. However, expanding our method becomes challenging in scenarios with heterogeneous noise matrices, e.g., the entries of noise matrices have different and unknown variances. If a uniform upper bound for the variances of noise exists, a comparable error bound can still be derived, similar to the results in least squares estimators. However, this bound might be inadequate, especially when the upper bounds for variances are significantly larger than the normalized spectral norm of the noise matrix. Therefore, it is meaningful to develop methods that adapt the proposed approach to more complex noise scenarios in the future.






\bibliographystyle{imsart-nameyear} 
\bibliography{bibliography}       

\end{document}